%% file: iclr2027_conference.tex
\documentclass{article} 
\usepackage{iclr2027_conference,times}

\input{math_commands.tex}

\usepackage{url}

\usepackage[T1]{fontenc}
\usepackage[utf8]{inputenc}
\usepackage{microtype}
\usepackage{amsmath,amssymb}
\usepackage{booktabs}
\usepackage{graphicx}
\usepackage{multirow}
\usepackage[table]{xcolor}
\usepackage{xcolor}
\usepackage{colortbl}

\definecolor{citecolor}{HTML}{0071bc}
\usepackage[pagebackref=false,breaklinks=true,letterpaper=true,colorlinks,citecolor=citecolor,bookmarks=false]{hyperref}

\usepackage{url}
\usepackage[capitalize,noabbrev]{cleveref}

\graphicspath{{../figures/}{figures/}}

\usepackage{amsthm}
\definecolor{resultwin}{RGB}{185,35,45}
\definecolor{resultnonwin}{RGB}{20,125,75}
\definecolor{tablegroupgeneral}{RGB}{244,245,248}
\definecolor{tablegroupaligned}{RGB}{237,243,249}
\definecolor{tablegrouplatent}{RGB}{238,246,242}
\definecolor{tablemethodhighlight}{RGB}{252,241,243}
\newcommand{\method}{\textsc{Latent-JEPA}}
\newcommand{\localwin}[1]{\textcolor{resultwin}{\textbf{#1}}}
\newcommand{\localtie}[1]{\textcolor{resultwin}{\textbf{#1}}}

\newcommand{\sg}{\operatorname{sg}}
\newcommand{\norm}[1]{\left\lVert #1 \right\rVert_2}
\newtheorem{definition}{Definition}

\title{\method{}: Abstract Future Prediction \\ for Latent Reasoning in Chemistry}
\author{Xinjian Zhao\textsuperscript{$\S$ \P },
Yaoyao Xu\textsuperscript{$\S$ \P }, 
Xuemin Chen\textsuperscript{$\S$ \P},
Xiaozhuang Song\textsuperscript{$\S$ \P}, 
Tianshu Yu\textsuperscript{$\S$ \P }
\\
\textsuperscript{$\S$} School of Data Science, The Chinese University of Hong Kong, Shenzhen \\
\textsuperscript{\P} Shanghai Artificial Intelligence Laboratory\\
 \texttt{\{xinjianzhao1,yaoyaoxu,xueminchen,xiaozhuangsong1\}@link.cuhk.edu.cn}, \\
 \texttt{yutianshu@cuhk.edu.cn} 
}

\iclrfinalcopy

\begin{document}
\maketitle
\begin{abstract}
Large language models offer a promising foundation for chemical reasoning,
bringing together chemical knowledge and multistep problem solving.
Chemical intuition can provide an initial sense of plausible outcomes
before the details of a solution are fully worked out.
Inspired by how such expectations complement explicit analysis, we study
how continuous latent thoughts can be trained to anticipate informative
aspects of future solutions without verbalizing every intermediate step. We introduce
\method{}, a framework that combines autoregressive learning with
joint-embedding prediction of one or more future views.  For chemical
reasoning, we develop textual and molecular prediction objectives that
connect latent thoughts to both subsequent reasoning and molecular outcomes.
Experiments on ChemCoTBench show gains in molecular
optimization and on several editing and reaction metrics. Representation
analyses show that future prediction makes latent thoughts more informative
about molecular outcomes and strengthens their correspondence with chemical
structure. These findings support abstract future prediction as a learning
principle for connecting continuous latent reasoning with scientific outcomes.
\end{abstract}

\section{Introduction}
\label{sec:introduction}

Scientific discovery often begins before a complete explanation is available. Researchers form hypotheses, recognize promising directions, and develop expectations about possible outcomes while important details remain unresolved. This interplay between provisional understanding and explicit analysis has long been discussed in accounts of scientific reasoning. \citet{poincare2003science} emphasized the complementary roles of intuition and logical demonstration in mathematical
invention, while \citet{polanyi2009tacit} argued that scientific
knowledge contains an important tacit component that cannot always be fully
articulated. Cognitive studies likewise suggest that
people can become sensitive to meaningful coherence before they can explicitly
identify it \citep{bowers1990intuition}.
Such expectations do not replace formal reasoning or empirical validation;
they help identify and organize possibilities that can subsequently be
worked out and tested. For scientific reasoning models~\citep{truhn2023large,narayanan2026training,pang2026paper2poster}, such anticipatory structure could complement the explicit multistep reasoning already used in scientific problem solving. We ask how a scientific reasoning model might learn
an analogous capability: forming useful expectations about a solution before
that solution is fully articulated.

In chemistry, intuition can provide an initial sense of plausible outcomes before the details of a solution are fully worked out. An experienced chemist
may, for example, have an initial sense of what transformations are plausible
or how a change in molecular structure might affect an outcome before working
through all of the details. Such judgments draw on accumulated chemical
knowledge and patterns learned through experience
\citep{talanquer2022complexity}. They complement explicit analysis and can be
refined and tested as reasoning proceeds. We draw inspiration from this
anticipatory aspect of chemical intuition to study how models can learn useful
expectations about future solutions.

Continuous latent reasoning provides a computational setting in which to
study this idea. Chain-of-thought (CoT) expresses intermediate reasoning in
language \citep{wei2022chain}, whereas continuous reasoning allows models to
update internal thoughts without verbalizing every step. Coconut implements
hidden-state feedback \citep{hao2024training}, CODI learns continuous reasoning
through self-distillation \citep{shen2025codi}, and LatentChem incorporates
dynamic molecular representations into chemical latent reasoning
\citep{ye2026latentchem}.
We ask a complementary learning question:
\emph{can predicting representations of future solutions improve both
the information retained in latent thoughts and the reasoning they support?}

An informative prediction need not specify every detail of what comes next.
When a glass tips over the edge of a table, one may anticipate that it could
break without predicting the position of every shard
\citep{battaglia2013simulation}.
Similarly, an improvising poet may be guided by an implicit sense of
rhyme without explicitly planning the line in advance. 
These examples motivate studying how internal representations can
anticipate meaningful aspects of a future before its full realization.
Joint-Embedding Predictive Architectures (JEPAs) offer a learning principle
aligned with this distinction: predict an encoded view rather than reconstruct
its observable details \citep{dawid2024introduction,assran2023self,bardes2024revisiting}.
Predictive video models have also shown sensitivity to violations of physical
expectations \citep{garrido2025intuitive}.
For continuous reasoning, this motivates training latent thoughts to predict
representations of subsequent reasoning and outcomes, while leaving their
explicit realization to autoregressive generation.

We introduce \method{}, which augments continuous latent reasoning with
joint-embedding prediction of future representations. \emph{The reasoner determines how latent computation unfolds, while future views specify what aspects of a developing solution it is trained to anticipate.} For chemical reasoning, we
instantiate this principle with two complementary views. A textual view
represents the remaining reasoning and answer in the language model's
representation space, while a molecular view uses a pretrained molecular
encoder to represent the chemical outcome \citep{soares2025smi}. The two views
capture different aspects of the same solution: different explanations can
describe the same molecular outcome, and similar forms of reasoning can lead
to different structures. A shared latent thought predicts both views in their
respective spaces, allowing linguistic and chemical outcomes to provide
complementary supervision without requiring a common target representation.

We evaluate \method{} on ChemCoTBench \citep{hao2026beyond}. \method{} improves molecular optimization and several editing and reaction metrics. Our analyses further show that predictive supervision makes individual molecular outcomes more recoverable from latent thoughts and brings recurrent representations into closer correspondence with molecular structure. Together, these results support abstract future prediction as a way to train latent thoughts to form useful expectations about future reasoning and chemical outcomes before explicit generation. More broadly, \method{} takes a step toward the anticipatory aspect of scientific intuition, where internal computation forms useful expectations about possible outcomes before an explicit solution is produced.

Our contributions are:
\begin{itemize}
    \item We introduce \method{}, a framework that augments continuous latent
    reasoning with prediction of one or more future views, separating the
    reasoning architecture from the choice of target representations.

    \item  We develop textual and molecular JEPA objectives that train latent thoughts to predict representations of subsequent reasoning and molecular outcomes, and demonstrate their complementary benefits for chemical reasoning.

    \item We show that future prediction makes latent thoughts more informative
    about molecular outcomes and strengthens their correspondence with
    molecular structure.
\end{itemize}

\section{Related Work}
\label{sec:related}
\noindent\textbf{Predictive representation learning.}
Predictive representation learning provides a way to supervise abstractions
of an observation. JEPA predicts embeddings of withheld views rather than
reconstructing their low-level details \citep{dawid2024introduction,assran2023self,bardes2024revisiting}.
Related non-contrastive methods study target asymmetry and representation
collapse \citep{grill2020bootstrap,bardes2021vicreg}, while LLM-JEPA combines
embedding prediction with language modeling \citep{huang2026llm}.
Future Summary Prediction and Semformer extend prediction to representations
of a future text segment \citep{mahajan2026beyond,yin2024semformer}.
We apply this principle to thoughts produced during continuous reasoning,
with language and molecular representations describing complementary futures.

\noindent\textbf{Continuous latent reasoning.}
Continuous reasoning changes how a model computes before answering. Coconut
feeds hidden states back into the language model and progressively replaces
explicit CoT with continuous thoughts \citep{hao2024training}; CODI and SoftCoT
develop related latent reasoning methods \citep{shen2025codi,xu2025softcot}.
Looped Transformers increase computation through repeated shared transformations
\citep{saunshi2025reasoning}.

\noindent\textbf{Chemical reasoning and representations.}
Chemical reasoning connects linguistic explanations with molecular structure
and task outcomes
\citep{narayanan2026training,wang2026chem,song2026aot,feng2026motifagent,zhu2026retroagent}.
LatentChem incorporates molecular representation updates into latent reasoning
\citep{ye2026latentchem}, while ChemCoTBench evaluates chemical transformations
and reasoning tasks \citep{hao2026beyond}. A complementary line of work learns
molecular representations from structure and text
\citep{chithrananda2020chemberta,ross2022large,liu2023multi,soares2025smi,zhao2026molemb}.
Together with language-model embeddings \citep{behnamghader2024llm2vec},
these representations provide ways to describe both a solution and its
molecular outcome. We use them as future prediction targets;
Appendix~\ref{app:additional-related} further connects target design with scientific representation learning.

\begin{figure*}[h]
    \centering
    \includegraphics[width=\textwidth]{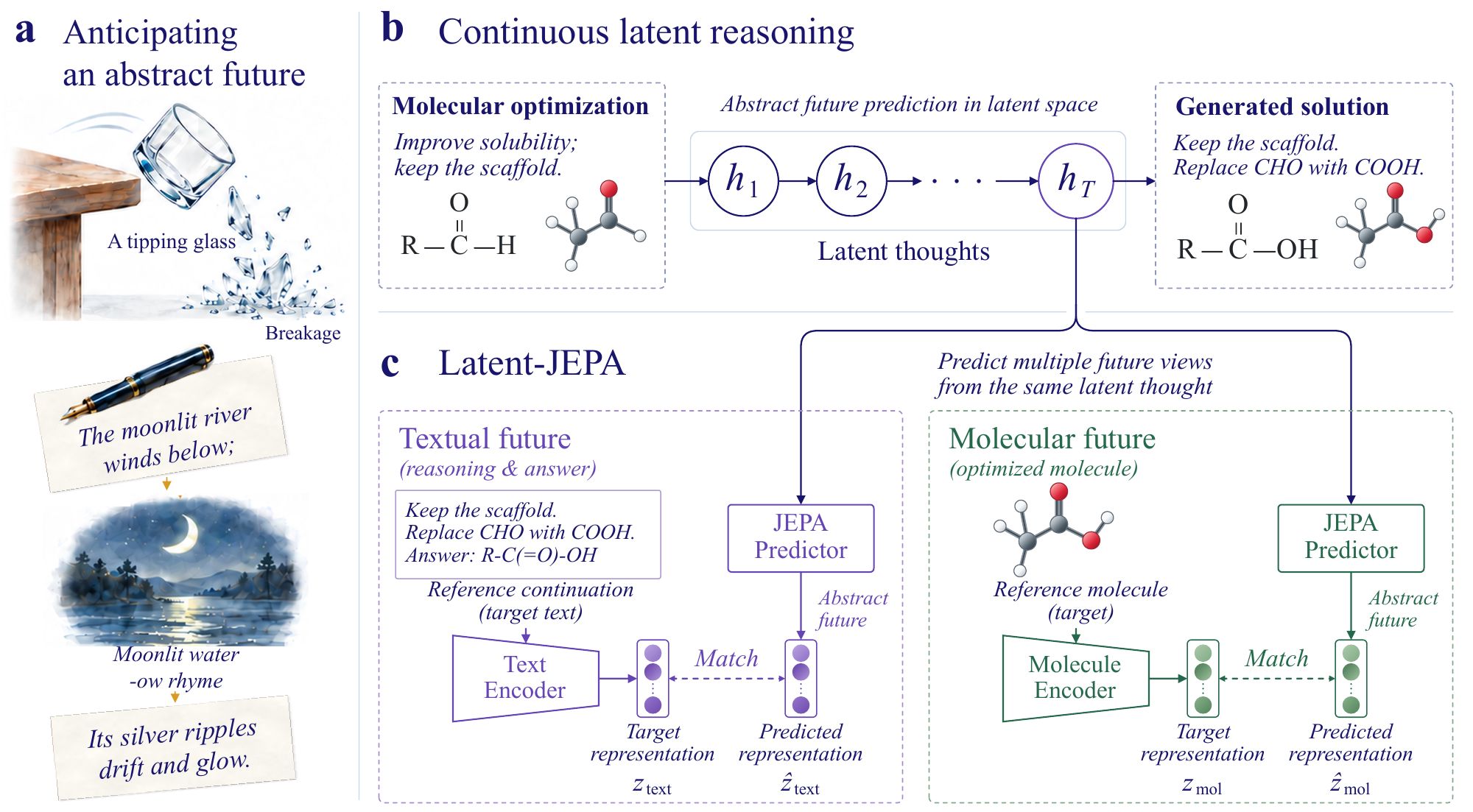}
    \caption{Overview of \method{}. 
    \textbf{(a)} Anticipating breakage or a poem's
    theme and rhyme illustrates prediction at the level of useful
    abstractions. The verse gives one possible continuation of the first line.
    \textbf{(b)} A continuous reasoner updates latent thoughts before
    generating the remaining reasoning and molecular outcome.
    \textbf{(c)} The final latent thought also predicts textual and molecular
    representations of the reference solution during training. 
    The molecular optimization
    illustration replaces an aldehyde with a carboxyl group; $R$ denotes the
    unchanged scaffold.}
    \label{fig:overview}
\end{figure*}
\section{Latent-JEPA}
\label{sec:method}

We aim to train latent thoughts to form useful expectations about a
solution before it is fully expressed. To this end, \method{} combines
autoregressive learning with prediction of representations of subsequent
reasoning and outcomes (\cref{fig:overview}). Let $W_t$ denote the reasoner's current workspace
and $h_t$ a continuous representation of intermediate reasoning read
from that workspace:
\begin{equation}
    W_0=I_\theta(x),\qquad h_t=R_\theta(W_t),\qquad
    W_{t+1}=U_\theta(W_t,h_t).
    \label{eq:latent-workspace}
\end{equation}
Here $x$ is the input, $I_\theta$ initializes the workspace, $R_\theta$ reads
the thought, and $U_\theta$ advances the reasoning process for $0\leq t<K$.
A decoder generates the answer from $W_K$. The workspace can include the
reasoner's attention cache or domain-specific state, allowing this formulation
to cover different continuous reasoning architectures.

\begin{definition}[Latent-JEPA]
\label{def:latent-jepa}
A Latent-JEPA architecture consists of a continuous latent reasoner and, at
each supervised step $t\in\mathcal I\subseteq\{0,\ldots,K\}$, a finite family
of future-view maps, target encoders, and predictors
$\{(\phi_{t,j},E_j,P_j)\}_{j=1}^{N_t}$, where $N_t\geq1$.
Given a reference continuation or outcome $F_{>t}$ available during
training, $\phi_{t,j}$ selects a future view, $E_j$ encodes it,
and $P_j$ predicts its representation from $h_t$:
\begin{equation}
    v_{t,j}=\phi_{t,j}(F_{>t}),\qquad
    z_{t,j}=E_j(v_{t,j}),\qquad
    \widehat z_{t,j}=P_j(h_t,q_{t,j}),
    \label{eq:generic-target}
\end{equation}
where $q_{t,j}$ optionally specifies a prediction horizon or query.
The latent thought $h_t$ depends only on computation preceding the future
view. Each predictor maps the thought into its target encoder's representation
space, where a discrepancy $d_j(\widehat z_{t,j},z_{t,j})$ supplies the
predictive learning signal.
\end{definition}

A future view can describe an intermediate reasoning step, the answer, or a
domain-specific outcome. Different views can use different encoders and
representation spaces; multiple horizons can also share an encoder and a
query-conditioned predictor. The framework leaves these choices open so that
new target representations and reasoning architectures can extend the same
predictive structure.
The definition also leaves the use of the components after training open.
Encoders and predictors may support downstream planning or evaluation using
information available at inference. In this work, we use the predictive branches to supervise latent thoughts
during training, without changing the reasoner's generation procedure.

\noindent\textbf{Predictive learning.}
We train the predictors and latent reasoner with
\begin{equation}
    \mathcal L_{\mathrm{LJ}}
    =\mathbb E\!\left[
       \sum_{t\in\mathcal I}\sum_{j=1}^{N_t}
       w_{t,j}\,
       d_j\!\left(P_j(h_t,q_{t,j}),
                  \sg(E_j(v_{t,j}))\right)
    \right],\qquad w_{t,j}\geq0,
    \label{eq:generic-jepa}
\end{equation}
where the expectation is over training examples and $w_{t,j}$ controls the
contribution of each view and reasoning step. Stop-gradient prevents the
prediction loss from directly updating the target branch. The encoder may
remain frozen, follow a momentum update, or share parameters with the reasoner;
these choices determine how the targets evolve during learning.
Target calibration, when used, forms part of the map $E_j$.
A simple choice is to supervise the final latent thought, which can
integrate information from preceding latent steps without accessing
the future continuation. We adopt this choice, setting
$\mathcal I=\{K_s\}$, and combine the predictive loss with the
existing training objective:
\begin{equation}
    \mathcal L_{\mathrm{adapt}}
    =\mathcal L_{\mathrm{AR}}+\mathcal L_{\mathrm{base}}
     +\lambda_{\mathrm{LJ}}\mathcal L_{\mathrm{LJ}}.
    \label{eq:generic-total}
\end{equation}
Here $\mathcal L_{\mathrm{base}}$ denotes any architecture-specific auxiliary
losses. The chemical construction below specifies the future views, encoders,
and normalization of their contributions.

\subsection{Target asymmetry and geometry}

The target branch observes information withheld from the latent thought.
Stop-gradient makes the prediction asymmetric: the predictor and reasoner
learn to approach the target representation without directly optimizing the
target through the same loss. When encoder parameters are shared with the
reasoner, they still evolve through autoregressive and prediction-side
updates. A frozen encoder instead provides a fixed representation space.
Target geometry also determines which prediction errors are informative.
If embeddings share a dominant common direction, high cosine similarity may
reveal little about the corresponding example. We center target representations
using means estimated from training data and apply $\ell_2$ normalization
to expose variation across examples. We evaluate
this geometry through pairwise similarity and effective rank, and test
sample-specific prediction using shuffled targets. These are diagnostic
controls; the training objective in \cref{eq:generic-jepa} is non-contrastive.

\section{Latent-JEPA for Chemical Reasoning}
\label{sec:setting}

We use Coconut-Chem, the chemical adaptation of Coconut introduced by
\citet{ye2026latentchem}, as a representative latent reasoner. We first
describe its latent computation, then define textual and molecular
prediction objectives. Additional instantiations of \method{} are given in
Appendices~\ref{app:latentchem-transfer} and~\ref{app:vanilla-transfer}.

\subsection{Recurrent chemical reasoning}

Let $x$ denote the input prompt and any observable molecular information.
A demonstrated solution consists of reasoning steps $c_{1:M}$ and a final
answer $y$, where $M$ is the number of explicit reasoning steps. Following
Coconut \citep{hao2024training}, Coconut-Chem progressively replaces an
increasing prefix of these steps with continuous thoughts. At curriculum
stage $s$, the first $s$ reasoning steps are replaced by a latent sequence
$h_{1:K_s}$, where $K_s$ denotes the latent reasoning length at that stage.
The model then autoregressively generates the remaining reasoning
$c_{s+1:M}$ followed by $y$.
For stages with $K_s>0$, we supervise the final latent thought,
\begin{equation}
    h_T \equiv h_{K_s},
    \label{eq:terminal-state}
\end{equation}
where $h_{K_s}$ is the final-layer hidden state at the last continuous-thought
position, before the end-of-reasoning delimiter. The causal attention mask
prevents this state from accessing the remaining reasoning or answer tokens.
The autoregressive objective is
\begin{equation}
    \mathcal{L}_{\mathrm{AR}}
    = - \sum_{t \in \mathcal{S}}
      \log p_\theta(w_t \mid x,h_{1:K_s},w_{<t}),
    \label{eq:ar}
\end{equation}
where $w_t$ is a continuation token and $\mathcal S$ indexes the supervised
tokens in $c_{s+1:M}$ and $y$. Gradients from \cref{eq:ar} train latent
computation through continuation prediction. \method{} additionally
supervises $h_T$ with the textual and molecular targets available for each
example. Thus, $\mathcal I=\{K_s\}$ at stages with continuous thoughts.
The curriculum schedule is specified in Appendix~\ref{app:contract}.

\subsection{Decoder-native continuation view}
\label{sec:reasoning-horizons}
The textual view trains the final latent thought to anticipate the
remaining solution in the generator's own representation space.
We construct these decoder-native targets with a separate causal
forward pass over the reference continuation $(c_{s+1:M},y)$.
This pass shares the current decoder parameters with the reasoner,
but receives only the continuation tokens, without the prompt,
source-molecule features, or latent thoughts.
We apply stop-gradient to the target representations.
From this single pass, we read the final-layer hidden state at the
last content token of each reasoning step and of the answer.
A shared predictor $P_{\mathrm{text}}(h_T,\rho)$ predicts these
representations from the same final latent thought, with $\rho$
specifying the relative horizon in the remaining solution.
We center the targets using separate running means for reasoning
and answers, then apply $\ell_2$ normalization.
The objective $\mathcal L_{\mathrm{reason}}$ minimizes cosine
distance $d(a,b)=1-\cos(a,b)$, assigning equal total weight to
reasoning and answer prediction when both are available.
Appendix~\ref{app:text-target-construction} gives the encoding
details, horizon queries, and loss normalization.

\subsection{Structure-native outcome view}
\label{sec:chemical-grounding}
The textual view trains latent thoughts to anticipate the continuation
in the language model's representation space. We complement this
supervision with a molecular view of the future outcome, using a
pretrained molecular encoder to provide a structural abstraction
of the solution.
For examples with a molecular outcome, let $m_i^{\mathrm{chem}}$
denote the reference molecule or molecular state and
$\tilde e_i = E_{\mathrm{SMI}}(m_i^{\mathrm{chem}})$ its representation
from a frozen SMI-TED-based encoder~\citep{soares2025smi}.
To make differences among outcomes more visible to the cosine
objective, we subtract the mean over molecular training targets
and apply $\ell_2$ normalization:
\begin{equation}
    z_i^{\mathrm{chem}}
    = \sg\left(
      \frac{\widetilde e_i-\mu_{\mathrm{train}}}
           {\norm{\widetilde e_i-\mu_{\mathrm{train}}}
           +\epsilon}
      \right).
    \label{eq:chem-target}
\end{equation}
The training mean $\mu_{\mathrm{train}}$ remains fixed. A separate predictor produces
$\widehat z_i^{\mathrm{chem}}=P_{\mathrm{chem}}(h_T)$, with
\begin{equation}
    \mathcal{L}_{\mathrm{chem}}^{(i)}
    =d(\widehat z_i^{\mathrm{chem}},z_i^{\mathrm{chem}}).
    \label{eq:chem-loss}
\end{equation}
Examples with text-only answers receive autoregressive and applicable
textual prediction supervision, but no molecular prediction objective.
Appendix~\ref{app:contract} describes the preparation of molecular targets.
The two predictors map the same latent thought into separate target spaces,
allowing textual and molecular representations to preserve different aspects
of the solution.

\subsection{Objective and inference}

We retain Coconut-Chem's autoregressive and counterfactual objectives
and add the two predictive losses:
\begin{equation}
\mathcal{L}
=\mathcal{L}_{\mathrm{AR}}
+\mathcal{L}_{\mathrm{CF}}
+\lambda_{\mathrm{reason}}\mathcal{L}_{\mathrm{reason}}
+\lambda_{\mathrm{chem}}\mathcal{L}_{\mathrm{chem}},
\label{eq:full-objective}
\end{equation}
where $\mathcal{L}_{\mathrm{CF}}$ is Coconut-Chem's molecular-input
counterfactual objective. Each predictive loss is averaged over its eligible
examples and uses unperturbed inputs; Appendix~\ref{app:contract} gives the details.

At inference, the model follows the underlying reasoner's recurrent
computation and decoding procedure. Prediction heads and target encoding
serve only as training supervision.

\section{Experiments}
\label{sec:experiments}

We first evaluate whether anticipating future solutions improves chemical
reasoning. We then examine what latent thoughts learn about molecular
outcomes and which target choices provide useful predictive supervision.

\subsection{Experimental Setup}

We implement Coconut-Chem with Qwen3-8B-Base \citep{yang2025qwen3}, initialized
from LatentChem's chemistry instruction-tuned model and retaining its SMI-TED
molecular input encoder and adapter. Coconut-Chem and its dual-view
\method{} variant share initialization, training data, and a curriculum
with $0,2,4,$ and $6$ continuous thoughts, trained for three epochs per phase.
We also evaluate answer prediction in LatentChem and compare future views
in Vanilla Coconut, which receives molecular SMILES as text without source
ChemTokens. Architecture-specific training settings are detailed in
Appendix~\ref{app:contract}.
We evaluate on ChemCoTBench \citep{hao2026beyond,ye2026latentchem}, which
covers molecular optimization, editing, molecular understanding, scaffold
abstraction, and forward-reaction prediction. We use the corresponding
property, accuracy, count-error, and structural-similarity metrics, including
fingerprint Tanimoto similarity (FTS) for reactions. Each checkpoint generates
three responses per problem at temperature $0.7$ and top-$p=0.9$; we average
their scores without best-of-three selection. Our within-architecture
comparisons share evaluation settings, while published baselines retain
their original protocols. Appendix~\ref{app:baseline-provenance} documents
baseline sources, and Appendix~\ref{app:integrity} specifies the tasks,
metrics, and scoring procedure.

\subsection{Main Results}
\label{sec:results}

\begin{table*}[t]
\centering
\small
\renewcommand{\arraystretch}{1.12}
\setlength{\tabcolsep}{3.5pt}
\caption{Molecule editing and optimization on ChemCoTBench. Editing reports
correctness (\%); optimization reports mean property improvement. All metrics
are higher-is-better. Dark-red bold marks
\method{} scores equal to or better than Coconut-Chem. Optimization
success rates appear in \cref{tab:optimization-success}.}
\label{tab:molopt-results}
\resizebox{\textwidth}{!}{%
\begin{tabular}{lrrrrrrrrr}
\toprule
& \multicolumn{3}{c}{Molecule editing (\%)} & \multicolumn{6}{c}{Property improvement} \\
\cmidrule(lr){2-4}\cmidrule(lr){5-10}
Method & Add & Delete & Sub. & LogP & Solubility & QED & DRD2 & JNK3 & GSK3-$\beta$ \\
\midrule
\rowcolor{tablegroupgeneral}
\multicolumn{10}{l}{\textit{General-purpose language models}} \\
GPT-4o & 80.0 & 80.0 & 65.0 & -0.09 & 0.92 & 0.13 & 0.07 & -0.02 & 0.00 \\
o1-mini & 55.0 & 80.0 & 58.3 & -0.42 & 1.78 & 0.07 & -0.03 & -0.10 & -0.08 \\
o3-mini & 65.0 & 55.0 & 80.0 & 0.26 & 0.81 & 0.21 & 0.19 & -0.03 & 0.01 \\
Gemini 2.5 Pro (thinking) & 100.0 & 85.0 & 81.7 & -0.22 & 1.06 & 0.28 & 0.36 & -0.02 & 0.06 \\
Claude 3.7 Sonnet (thinking) & 85.0 & 80.0 & 83.4 & 0.41 & 0.37 & 0.12 & 0.17 & 0.01 & 0.02 \\
\midrule
\rowcolor{tablegroupaligned}
\multicolumn{10}{l}{\textit{Open and chemistry-aligned language models}} \\
Qwen3-8B & 4.0 & 52.0 & 1.0 & 0.00 & 0.00 & 0.00 & 0.00 & -- & -- \\
Qwen3-8B (SFT) & 55.0 & 76.0 & 37.0 & 0.15 & 0.48 & 0.10 & 0.04 & -- & -- \\
Molecular alignment & 62.0 & 61.0 & 47.0 & 0.35 & 0.87 & 0.14 & 0.20 & -- & -- \\
Molecular alignment + CoT & 63.0 & 80.0 & 49.0 & 0.51 & 0.82 & 0.14 & 0.12 & -- & -- \\
Molecular alignment + CoT + GRPO & 70.0 & 88.0 & 60.0 & 0.67 & 1.17 & 0.19 & 0.19 & -- & -- \\
\midrule
\rowcolor{tablegrouplatent}
\multicolumn{10}{l}{\textit{Latent reasoning models}} \\
LatentChem & 65.0 & 78.3 & 56.7 & 0.7051 & 0.7829 & 0.1809 & 0.1732 & 0.0358 & 0.1481 \\
\textbf{Coconut-Chem} & 71.7 & 76.7 & 65.6 & 0.7404 & 0.9259 & 0.2181 & 0.2636 & 0.1032 & 0.2285 \\
\rowcolor{tablemethodhighlight}
\textbf{Coconut-Chem + \method{}}
& \localwin{78.3} & \localwin{86.7} & 61.1
& \localwin{0.8318} & \localwin{0.9659} & \localwin{0.2296}
& \localwin{0.2848} & \localwin{0.1296} & \localwin{0.2515} \\
\bottomrule
\end{tabular}}\vspace{-3.9mm}
\end{table*}

\begin{table*}[t]
\centering
\small
\renewcommand{\arraystretch}{1.12}
\setlength{\tabcolsep}{3.5pt}
\caption{Molecular understanding and forward-reaction prediction. FG and Ring
report count MAE; Murcko and FTS report similarity; ring-system accuracy and
reaction validity are percentages. Dark-red bold marks \method{} scores equal to
or better than Coconut-Chem. Ring-system scoring and the treatment of
published reference scores are specified in Appendix~\ref{app:evaluation-scoring}.}
\label{tab:structure-reaction-results}
\resizebox{\textwidth}{!}{%
\begin{tabular}{lrrrrrrrr}
\toprule
& \multicolumn{4}{c}{Molecular understanding}
& \multicolumn{2}{c}{Major product}
& \multicolumn{2}{c}{Byproduct} \\
\cmidrule(lr){2-5}\cmidrule(lr){6-7}\cmidrule(lr){8-9}
Method & FG $\downarrow$ & Ring $\downarrow$ & Murcko $\uparrow$ & Ring-sys. (\%) $\uparrow$
& FTS $\uparrow$ & Valid (\%) $\uparrow$ & FTS $\uparrow$ & Valid (\%) $\uparrow$ \\
\midrule
\rowcolor{tablegroupgeneral}
\multicolumn{9}{l}{\textit{General-purpose language models}} \\
GPT-4o & 0.17 & 1.35 & 0.21 & 80.0 & 0.58 & -- & 0.20 & -- \\
o1-mini & 0.21 & 1.25 & 0.25 & 61.7 & 0.31 & -- & 0.17 & -- \\
o3-mini & 0.13 & 0.60 & 0.39 & 75.0 & 0.71 & -- & 0.27 & -- \\
Gemini 2.5 Pro (thinking) & 0.11 & 0.60 & 0.51 & 87.5 & 0.89 & -- & 0.51 & -- \\
Claude 3.7 Sonnet (thinking) & 0.21 & 1.60 & 0.40 & 80.0 & 0.87 & -- & 0.31 & -- \\
\midrule
\rowcolor{tablegroupaligned}
\multicolumn{9}{l}{\textit{Open and chemistry-aligned language models}} \\
Qwen3-8B & 7.03 & 1.03 & 0.02 & 36.0 & 0.02 & -- & 0.00 & -- \\
Qwen3-8B (SFT) & 0.12 & 0.45 & 0.40 & 65.0 & 0.35 & -- & 0.07 & -- \\
Molecular alignment & 0.14 & 0.10 & 0.71 & 83.0 & 0.57 & -- & 0.18 & -- \\
Molecular alignment + CoT & 0.08 & 0.20 & 0.66 & 63.0 & 0.44 & -- & 0.11 & -- \\
Molecular alignment + CoT + GRPO & 0.07 & 0.15 & 0.65 & 83.0 & 0.63 & -- & 0.20 & -- \\
\midrule
\rowcolor{tablegrouplatent}
\multicolumn{9}{l}{\textit{Latent reasoning models}} \\
LatentChem & 0.0967 & 0.1667 & 0.7220 & 73.89 & 0.5099 & 74.00 & 0.1728 & 42.67 \\
\textbf{Coconut-Chem} & 0.0667 & 0.1500 & 0.7783 & 80.00 & 0.6408 & 86.67 & 0.1518 & 45.33 \\
\rowcolor{tablemethodhighlight}
\textbf{Coconut-Chem + \method{}}
& \localtie{0.0667} & \localwin{0.1333} & \localwin{0.8065} & \localwin{81.11}
& \localwin{0.6492} & \localwin{88.33} & \localwin{0.1806} & \localtie{45.33} \\
\bottomrule
\end{tabular}}\vspace{-3.9mm}
\end{table*}

\noindent\textbf{Future prediction improves chemical reasoning.}
We first test the central premise of \method{} by comparing chemical task
performance with and without future prediction.
\Cref{tab:molopt-results,tab:structure-reaction-results} show improvements
on most reported metrics. The gains are most consistent in molecular
optimization, where mean property improvement increases on all six
objectives. Improvements also extend to structural analysis and reaction
prediction, while editing results are more mixed. The separate optimization
success rates improve on most objectives
(\cref{tab:optimization-success} in Appendix~\ref{app:optimization-success}).
Together, these findings support training latent thoughts to anticipate
abstract outcomes as a useful complement to autoregressive learning for
chemical reasoning.

\noindent\textbf{The two future views contribute differently.}
A solution's textual continuation and molecular outcome describe different
aspects of its future. We therefore compare textual-only, molecular-only,
and dual prediction to understand the contribution of each view
(\cref{tab:prediction-view-performance} in
Appendix~\ref{app:prediction-view-ablation}). Molecular prediction is
particularly effective for editing, whereas adding textual prediction
improves most property objectives and the reaction metrics relative to
molecular prediction alone. The combined objective thus extends the benefits
across a broader range of chemical tasks without being best on every metric.
These results motivate choosing predictive views according to the aspects
of a solution that a task requires; the loss coefficients for each comparison
are specified with the full results.

\noindent\textbf{Future prediction extends beyond Coconut-Chem.}
To examine whether the benefits depend on Coconut-Chem's particular reasoning
setup, we also apply the objective to LatentChem and Vanilla Coconut.
In LatentChem, predicting the answer representation improves most evaluated
metrics under the same supervised adaptation budget
(\cref{tab:latentchem-transfer} in Appendix~\ref{app:latentchem-transfer}).
In Vanilla Coconut, which has no source molecular encoder, textual
prediction gives the strongest average rank, while the other views benefit
different subsets of tasks
(\cref{tab:vanilla-prediction-view-performance} in
Appendix~\ref{app:prediction-view-ablation}). Future prediction can therefore
support more than one reasoning setup, with different setups favoring
different predictive views.

\subsection{Learning Predictive Latent Thoughts}
\label{sec:discussion}

To understand what accompanies the task gains, we examine whether latent
thoughts distinguish individual molecular outcomes, reflect their structural
relationships, and change through recurrent computation.

\begin{figure*}[t]
    \centering
    \includegraphics[width=\textwidth]{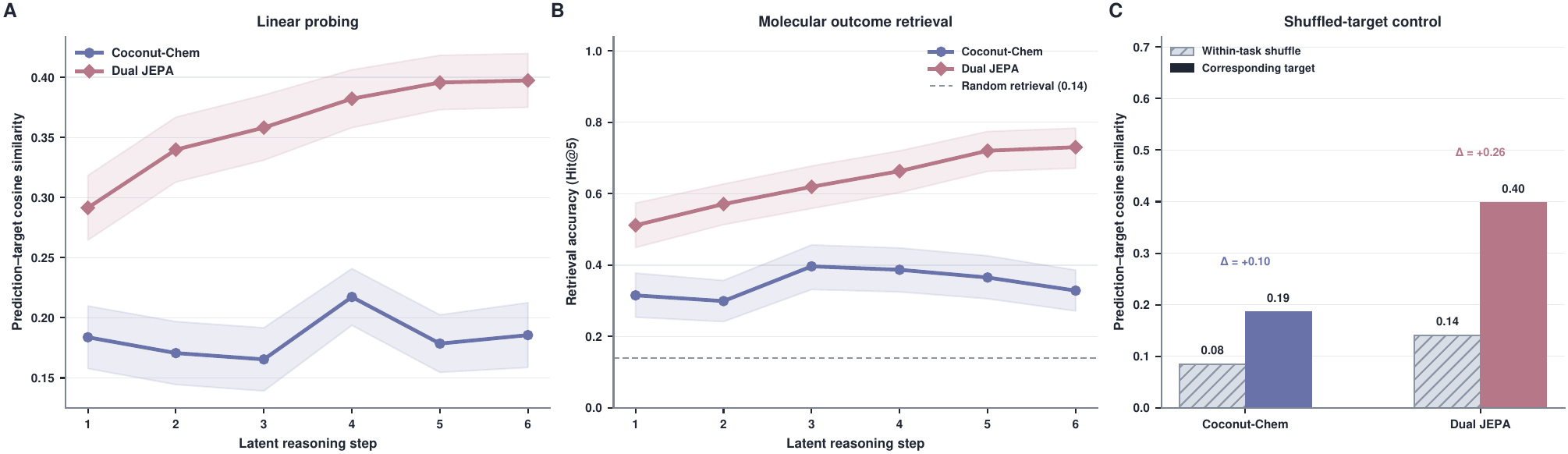}\vspace{-3.9mm}
\caption{Molecular-outcome prediction on examples excluded from reasoner training and probe fitting.
    \textbf{A:} Ridge-probe cosine similarity across recurrent updates.
    \textbf{B:} Within-task Hit@5; the dashed line denotes uniform retrieval.
    \textbf{C:} Corresponding-target similarity versus within-task shuffled targets.
     Bands summarize
    evaluation-example uncertainty; Appendix~\ref{app:analysis-methods}
    specifies the protocol.}

    \label{fig:heldout-decodability}
\end{figure*}

\noindent\textbf{Latent thoughts distinguish molecular outcomes.}
Anticipating an outcome requires more than recognizing the type of problem.
We therefore test whether a pre-answer thought contains information about
its particular molecular outcome. We fit a separate ridge probe to each
model and recurrent depth, then evaluate predictions on examples excluded from both reasoner training and probe fitting (Appendix~\ref{app:analysis-methods}).
\Cref{fig:heldout-decodability} shows higher prediction cosine and
within-task retrieval with \method{}; final-update cosine increases from
$0.186$ to $0.397$. Predictions also remain more similar to their corresponding
targets than to targets shuffled within the same task. This comparison
indicates that the improvement concerns individual outcomes, rather than
task membership alone. The example-wise and task-wise analyses show that
these gains extend across most analyzed examples and multiple task families
(\cref{fig:terminal-readout-scatter,fig:task-depth-decodability} in
Appendix~\ref{app:predictive-diagnostics}).

\begin{figure*}[t]
    \centering
    \includegraphics[width=\textwidth]{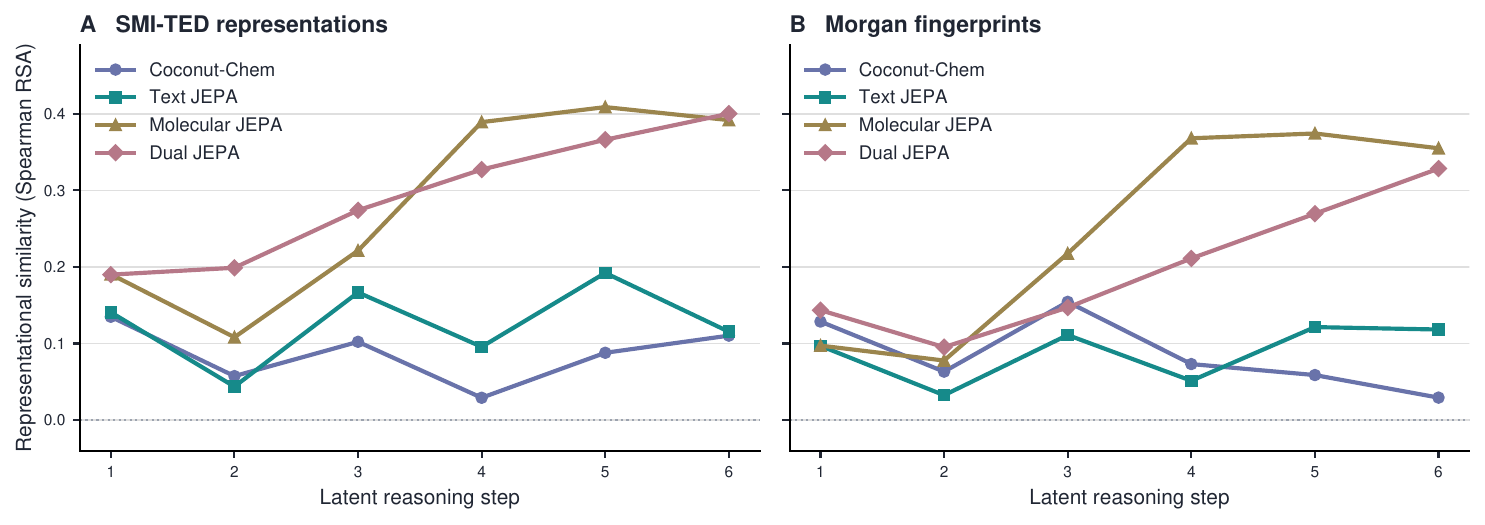}\vspace{-3.9mm}
    \caption{Representational similarity between recurrent feedback
    embeddings and molecular outcomes. \textbf{A:} Agreement with calibrated
    SMI-TED target distances. \textbf{B:} Agreement with Morgan fingerprint
    Tanimoto distances. Curves are task-averaged within-task Spearman RSA and
    single-run point estimates. Appendix~\ref{app:analysis-methods} specifies
    representations, configurations, and distance measures.}\vspace{-3.9mm}
    \label{fig:chemical-geometry}
\end{figure*}

\noindent\textbf{Recurrent representations reflect molecular relationships.}
Recovering an outcome with a learned probe does not reveal how the recurrent representations themselves are organized. We therefore compare pairwise distances between recurrent feedback embeddings and molecular outcomes using within-task representational similarity analysis (RSA), without fitting a mapping (Appendix~\ref{app:rsa-methods}).
In \cref{fig:chemical-geometry}, molecular-only and dual prediction show
substantially stronger agreement with SMI-TED geometry than Coconut-Chem or textual prediction. Molecular-only prediction also gives the strongest
agreement with Morgan fingerprints, which are not used as prediction targets. These results suggest that structure-native future targets can shape recurrent representations toward molecular relationships that extend beyond the target space itself. More generally, the choice of JEPA target determines which aspects of a future solution the latent state is encouraged to emphasize.
To examine the contribution of the molecular view more directly, we compare
final thoughts while holding each active view's coefficient fixed.
In this training-cohort analysis, molecular-only prediction gives the highest
outcome retrieval and dual prediction the strongest agreement with molecular
task-centroid relations (\cref{fig:terminal-latent-organization} in
Appendix~\ref{app:terminal-organization}). Molecular and dual prediction
also improve retrieval in Vanilla Coconut
(\cref{fig:vanilla-terminal-latent-organization}). These results motivate
using scientific encoders to supervise future outcomes, alongside their role
in encoding model inputs.

\noindent\textbf{Outcome representations develop through recurrent computation.}
We further examine how these representations change as latent reasoning
proceeds. The dual model's probe cosine and Hit@5 generally increase with
depth, with smaller changes near the final updates
(\cref{fig:heldout-decodability}). Its structural agreement also increases
overall (\cref{fig:chemical-geometry}), while the task-wise RSA gains follow
different depth profiles
(\cref{fig:chemical-rsa-heatmap} in Appendix~\ref{app:predictive-diagnostics}).
We also examine the explicit continuations and find task-dependent length
changes rather than a general shortening
(\cref{fig:residual-reasoning} in Appendix~\ref{app:residual-reasoning}).

\noindent\textbf{Constructing informative prediction targets.}
The preceding analyses concern what latent thoughts learn. We finally
examine how the target construction affects this supervision. A useful
predictive objective should distinguish different outcomes, rather than
reward a component shared by nearly every target. We inspect the molecular
target geometry and compare predictions using original and shuffled latent
sources. We find that raw SMI-TED targets share a dominant direction, so
uncentered targets can yield small prediction distances for both source
conditions. Centering and normalization make differences among outcomes
more visible to the cosine objective (\cref{fig:target-calibration} in
Appendix~\ref{app:training-target-diagnostics}). Retaining both centers
also gives the strongest average rank in the task ablation, with improvements
on most metrics over either removal (\cref{tab:centering-molopt} and the
complete results in \cref{tab:centering-all-tasks}). These observations
support considering target geometry together with the prediction loss when
constructing scientific supervision.

\begin{table}[t]
\centering
\small
\caption{Effect of target centering on molecular optimization. All entries
are mean property improvements ($\uparrow$).
Full task results are reported in \cref{tab:centering-all-tasks}.}
\label{tab:centering-molopt}
\begin{tabular}{lrrrrrr}
\toprule
Target calibration & LogP & Solubility & QED & DRD2 & JNK3 & GSK3-$\beta$ \\
\midrule
Both views centered & 0.8318 & 0.9659 & 0.2296 & 0.2848 & 0.1296 & 0.2515 \\
Without molecular centering & 0.8701 & 0.9703 & 0.2151 & 0.2645 & 0.1163 & 0.2329 \\
Without text centering & 0.7854 & 0.8311 & 0.2036 & 0.2644 & 0.1174 & 0.2232 \\
\bottomrule
\end{tabular}\vspace{-2.7mm}
\end{table}

The textual targets raise a related design question because their encoder
shares parameters with the reasoner and changes during training. We examine
whether a momentum encoder or a change in target dropout offers better
supervision, retaining both predictive views in each variant
(\cref{tab:target-encoder-ablation} in
Appendix~\ref{app:target-encoder-ablation}). Each variant improves a majority
of the reported metrics over Coconut-Chem, and the online encoder with
dropout has the strongest average rank among them. Thus, useful predictive
supervision is not confined to one target-update recipe. The shared online
encoder provides an effective source of textual futures, complementing the
fixed molecular representation. Together with the target-space analysis in
Appendix~\ref{app:target-structure}, these findings motivate studying which
scientific representations provide informative targets for reasoning.

\noindent\textbf{Discussion.}
Our results suggest a computational perspective on intuitive reasoning in continuous latent models. Scientific intuition often involves forming an expectation of a plausible outcome before the details of a solution are fully worked out; LATENT-JEPA instantiates this idea by training latent computation to anticipate abstract representations of future reasoning and outcomes before explicit generation. The representation analyses show that such supervision makes molecular outcomes more accessible from latent thoughts and reshapes recurrent computation toward chemically meaningful structure, suggesting that intermediate states can be trained not only to support eventual token generation, but also to encode useful expectations about what comes next. From this perspective, the choice of prediction target determines which aspects of the future are emphasized, while the predictive objective provides a more general mechanism for shaping the internal computation that precedes generation. More broadly, future prediction may influence not only what information is present in individual latent thoughts, but also how latent thoughts evolve through recurrent computation and how the underlying latent manifold is organized around possible future outcomes. Understanding these dynamics may provide a useful path toward intuitive reasoning in scientific models, where useful expectations emerge before a solution is explicitly worked out. Future work can further study how such latent trajectories develop across reasoning depth and whether different stages of computation benefit from different prediction horizons or levels of abstraction.

\section{Conclusion}

We introduced \method, a framework for learning latent thoughts through
prediction of future representations. Its formulation connects a continuous
reasoner to one or more future views and allows different encoders and
predictors to define the supervision. In chemistry, we train latent thoughts
to anticipate both the remaining reasoning and molecular outcomes.
Experiments show gains in chemical task performance and stronger molecular
information in latent thoughts. These findings motivate future prediction as
a learning objective for continuous reasoning and open further study of how
future representations can guide reasoning and planning.

\clearpage
\bibliographystyle{iclr2027_conference}
\bibliography{iclr2027_conference}
\clearpage
\appendix

\section{Additional Related Work}
\label{app:additional-related}

\noindent\textbf{Predictive targets for JEPA.}
When the prediction target is a representation rather than the observation
itself, constructing that target becomes part of the learning problem.
The encoder's available context and the level of detail it represents
help define what must be predicted. I-JEPA examines target scale and
contextual information when predicting representations of image regions
\citep{assran2023self}. V-JEPA studies feature prediction in video
\citep{bardes2024revisiting}, while data2vec predicts contextualized
representations from masked inputs across speech, vision, and language
\citep{baevski2022data2vec}. These studies provide a basis for considering
what the target encoder observes and how it represents that information.
Its update rule, including momentum updates, also determines how the target
space evolves during training. For chemical reasoning, we construct targets
from two views of a future solution: its textual continuation and the
molecular object it denotes. The encoder-update and calibration comparisons
in Appendix~\ref{app:target-encoder-ablation}, \ref{app:training-target-diagnostics} examine
how these design choices affect predictive supervision in our setting.

\noindent\textbf{Scientific and multimodal representations.}
Scientific representation models offer several ways to describe a molecular
outcome. MoleculeSTM relates structures to textual descriptions
\citep{liu2023multi}, while SMI-TED learns from molecular strings
\citep{soares2025smi}. Their structural and semantic representations
motivate using scientific encoders to define prediction targets, alongside
encoding observed inputs. In our molecular branch, a frozen SMI-TED-based
encoder represents the reference outcome during training; the language model
supplies the textual view of the remaining solution. Separate predictors map
the same latent thought to the two spaces. More broadly, this use of domain
encoders raises the question of which aspects of a scientific outcome provide
useful supervision for a reasoning task. Comparing encoders of the same
outcome examines the choice of representation, while selecting different
outcome views examines what information to predict.

\section{Implementation Details}
\label{app:contract}

\textbf{Optimization and hyperparameters.}
\phantomsection\label{app:training-settings}
Coconut-Chem uses AdamW with learning rate $2\times10^{-4}$, eight GPUs, four
examples per device, and no gradient accumulation. Each curriculum phase is
trained for three epochs. We use
$\lambda_{\mathrm{reason}}=0.025$ and $\lambda_{\mathrm{chem}}=0.01$,
with a decoding temperature of $0.7$ and top-$p=0.9$.
The loss-coefficient comparisons vary
$\lambda_{\mathrm{reason}}\in\{0.025,0.05,0.075\}$ and
$\lambda_{\mathrm{chem}}\in\{0.005,0.01,0.02\}$ one factor at a time around
$(0.05,0.01)$.
Text-predictor hidden widths of 512 and 1536 were also compared with the
912-dimensional reference. These comparisons use the benchmark
evaluation protocol described in Appendix~\ref{app:integrity}.
The representation-analysis configurations are specified in
\cref{app:analysis-methods}, including the text coefficient of $0.05$
used for the task-organization and recurrent-geometry comparisons.

\textbf{Curriculum.}
The reported curriculum allocates $c_{\mathrm{thought}}=2$ continuous
thoughts per replaced reasoning step and uses latent lengths
$K_s\in\{0,2,4,6\}$. Each phase is trained for three epochs. The baseline
and its \method{} variant use the same schedule.

\textbf{Supervised latent state.}
The predictor reads the final-layer hidden state at the last continuous-thought
position, before the end-of-reasoning delimiter. The initial,
fully textual curriculum phase has no latent thoughts and receives neither
predictive loss.

\textbf{Initialization and reference results.}
\phantomsection\label{app:baseline-provenance}
Coconut-Chem starts from LatentChem's chemistry-aware CoT model, called Stage~2 in
its training procedure. LatentChem adaptation starts after latent-module
training (its Stage~3) and precedes GRPO. Our adapted baseline and predictive
model receive the same additional supervised training. The
answer-prediction coefficient is $0.05$, and this transfer experiment has no
molecular predictive loss. We update the latent reasoning module and decoder
adapters while keeping the molecular updater frozen. Published values in
\cref{tab:molopt-results,tab:structure-reaction-results} use the ChemCoTBench
tables of \citet{hao2026beyond} for GPT-4o (2024-11-20), o1-mini
(2024-09-12), o3-mini (the source labels it 2025-01-03), Gemini 2.5 Pro
with thinking, and Claude 3.7 Sonnet with thinking. We retain their original
evaluation settings. Closed-model editing scores in ChemCoTBench use Pass@1.
For our models, we average scores from three sampled responses per problem,
evaluating each response separately as described in Appendix~\ref{app:complete-results}.
The Qwen and alignment rows come from the fifth arXiv revision of
\citet{ye2026latentchem}. Molecular alignment,
molecular alignment + CoT, and molecular alignment + CoT + GRPO correspond to
that paper's Stage~1, Stage~1+2, and Stage~1+2+4, respectively. Their original
evaluation protocol differs from our reproduced comparisons.

\textbf{Target visibility.}
The textual prediction branch encodes only the independently tokenized residual CoT
and wrapped answer. It cannot attend to the prompt, source molecules, or latent
states. Target hidden states are detached before centering and loss
computation. The frozen SMI-TED-derived encoder reads only the reference
chemical target (a molecule or scaffold-like structure, as applicable).

\textbf{Textual target construction and loss.}
\label{app:text-target-construction}
At curriculum stage $s$, the independent target sequence contains the
remaining reasoning, the answer enclosed by its answer delimiters, and an
end-of-sequence token:

\begin{equation}
    f^{\mathrm{text}}
    = [c_{s+1},\ldots,c_M,\langle\mathrm{answer}\rangle,
       y,\langle/\mathrm{answer}\rangle,\mathrm{EOS}].
\end{equation}

We encode this sequence in a separate causal forward pass using
the current decoder parameters, with gradients disabled and dropout
enabled by default. The pass does not reuse the reasoner's attention
cache. Each reasoning-step representation therefore depends only
on the continuation prefix through that step, while the answer
representation also includes the preceding residual reasoning.
We read the final-layer hidden representation at the
last semantic content token of each reasoning step and the last answer-content
token, excluding separators and delimiters. With $R=M-s$ remaining reasoning
steps, the predictor queries are $\rho_j=j/(R+1)$ for $j=1,\ldots,R$ and
$\rho_A=1$ for the answer. All predictions use the same final latent thought,
rather than supervising different latent depths.

For each representation $e^{\mathrm{text}}_j$, we subtract the running mean
for its target population and apply $\ell_2$ normalization:

\begin{equation}
    z^{\mathrm{text}}_{j}
    = \sg\left(
       \frac{e^{\mathrm{text}}_j-c^{\mathrm{text}}_{g(j)}}
       {\norm{e^{\mathrm{text}}_j-c^{\mathrm{text}}_{g(j)}}+\epsilon}
      \right),
    \label{eq:text-target}
\end{equation}

where $g(j)$ distinguishes reasoning-step and answer representations.
Center initialization, distributed aggregation, and update timing are
specified below.
With cosine distance $d(a,b)=1-\cos(a,b)$, the loss for example $i$ is:

\begin{equation}
\mathcal{L}_{\mathrm{reason}}^{(i)} =
\begin{cases}
\frac{1}{2}d(\widehat z_A,z_A)
+\frac{1}{2R}\sum_{j=1}^{R}d(\widehat z_j,z_j), & R>0,\\[3pt]
d(\widehat z_A,z_A), & R=0.
\end{cases}
\label{eq:reasoning-loss}
\end{equation}

This normalization gives every example a fixed total weight regardless of
CoT length; examples without remaining reasoning receive answer prediction
alone.

\textbf{Molecular target preparation.}
\phantomsection\label{app:molecular-target-preparation}
We encode the reference SMILES as provided, preserving component order.
The frozen SMI-TED-light encoder produces contextual token representations,
which we average over non-padding positions to obtain each molecular target.
We precompute these representations, subtract the mean over molecular
training targets, and apply $\ell_2$ normalization. The resulting targets
and training mean remain fixed during training.

\textbf{Running target means.}
We maintain separate running centers for reasoning and answer targets,
with EMA coefficient $\beta=0.99$. For reasoning targets, we first average
step representations within each example and then average across eligible
examples; answer targets are averaged across eligible examples. These
statistics are aggregated across distributed workers using sums and counts.
Both centers are computed only from training targets and reset at each
curriculum phase. The first eligible batch initializes each center from its
globally aggregated mean and uses that center immediately.
For subsequent batches, let $\bar e_{b,g}$ denote the globally aggregated
batch mean for target population $g$. The update is
\begin{equation}
    c_{b,g}=\beta c_{b-1,g}+(1-\beta)\bar e_{b,g},
\end{equation}
with the current batch using the pre-update center $c_{b-1,g}$ for target
normalization and the updated center used for later batches.
A center is unchanged when its target population has no eligible examples.
This EMA updates target means, not encoder parameters.
The molecular branch uses the fixed training-cache mean $\mu_{\mathrm{train}}$.

\textbf{Distributed reduction.}
We first average each prediction objective within an example, then average
over the examples eligible for that objective. Global sums and eligible-example
counts determine the normalization across distributed workers.

\textbf{Counterfactual objective.}
The counterfactual objective corrupts observable molecular features and is
enabled only when such features exist. Predictive losses use the unperturbed
input. Examples without molecular inputs receive no molecular counterfactual
loss.

\IfFileExists{latentchem_transfer_appendix.tex}{%
  \input{latentchem_transfer_appendix.tex}}{%
  \input{paper/iclr2027/latentchem_transfer_appendix.tex}}
  
\section{Target Encoder Ablation}
\label{app:target-encoder-ablation}
\label{sec:target-encoder-ablation}

We examine stochasticity and parameter updates in the text-target encoder.
The default shares the current language-model parameters and retains dropout.
The alternatives disable target dropout or use an exponential moving average
(EMA) of the LoRA parameters with momentum $0.995$ and dropout disabled.
All variants retain the molecular branch, predictors, loss coefficients,
and training schedule; generation always uses the current parameters.
The online-with-dropout configuration corresponds to the main model.
Average ranks are computed within this table over its 17 metrics using
unrounded scores and average ranks for ties; they summarize ordering, not
effect size.

\begin{table*}[t]
\centering
\small
\renewcommand{\arraystretch}{1.1}
\setlength{\tabcolsep}{3pt}
\caption{Text-target encoder ablation. Editing, ring-system
accuracy, and validity are percentages; property scores are mean improvements;
counts are MAE; Murcko and FTS are similarities. Dark-red bold marks the best
task score in each column, including ties at the displayed precision. Shading
identifies the default configuration. Lower average rank is better.}
\label{tab:target-encoder-ablation}
\resizebox{\textwidth}{!}{%
\begin{tabular}{lrrrrrrrrr}
\toprule
& \multicolumn{3}{c}{Molecule editing (\%) $\uparrow$} & \multicolumn{6}{c}{Property improvement $\uparrow$} \\
\cmidrule(lr){2-4}\cmidrule(lr){5-10}
Text target & Add & Delete & Sub. & LogP & Solubility & QED & DRD2 & JNK3 & GSK3-$\beta$ \\
\midrule
\rowcolor{tablemethodhighlight}
\textbf{Online, dropout} & 78.3 & \localwin{86.7} & 61.1 & 0.8318 & \localwin{0.9659} & 0.2296 & 0.2848 & \localwin{0.1296} & \localwin{0.2515} \\
Online, no dropout & 80.0 & 80.0 & 56.1 & \localwin{0.8613} & 0.9398 & 0.2168 & 0.2705 & 0.1104 & 0.2222 \\
EMA, no dropout & \localwin{81.7} & 80.0 & \localwin{64.4} & 0.8410 & 0.9549 & \localwin{0.2308} & \localwin{0.2957} & 0.1295 & 0.2278 \\
\bottomrule
\end{tabular}
}
\par\vspace{1.4em}\noindent
\resizebox{\textwidth}{!}{%
\begin{tabular}{lrrrrrrrr}
\toprule
& \multicolumn{4}{c}{Molecular understanding} & \multicolumn{2}{c}{Major product} & \multicolumn{2}{c}{Byproduct} \\
\cmidrule(lr){2-5}\cmidrule(lr){6-7}\cmidrule(lr){8-9}
Text target & FG $\downarrow$ & Ring $\downarrow$ & Murcko $\uparrow$ & Ring-sys. (\%) $\uparrow$ & FTS $\uparrow$ & Valid (\%) $\uparrow$ & FTS $\uparrow$ & Valid (\%) $\uparrow$ \\
\midrule
\rowcolor{tablemethodhighlight}
\textbf{Online, dropout} & \localwin{0.0667} & \localwin{0.1333} & \localwin{0.8065} & 81.11 & \localwin{0.6492} & \localwin{88.33} & 0.1806 & 45.33 \\
Online, no dropout & 0.0767 & 0.1500 & 0.7515 & \localwin{81.67} & 0.6357 & 87.33 & 0.1765 & \localwin{47.00} \\
EMA, no dropout & 0.0800 & 0.1500 & 0.7793 & 81.11 & 0.6147 & 85.33 & \localwin{0.1823} & 46.00 \\
\bottomrule
\end{tabular}
}
\par\vspace{0.8em}\noindent
\begin{tabular}{lr}
\toprule
Text target & Avg. rank $\downarrow$ \\
\midrule
Online, dropout & \textbf{1.68} \\
Online, no dropout & 2.35 \\
EMA, no dropout & 1.97 \\
\bottomrule
\end{tabular}
\end{table*}

The variants favor different tasks. Disabling dropout improves LogP but
reduces deletion and substitution accuracy. EMA improves substitution, QED,
DRD2 over the default, while reducing deletion
accuracy and major-product prediction. The online encoder thus provides
useful targets in this setting, without a uniform advantage from momentum
updates. These results motivate considering target updates together with
the information represented by the encoder, rather than treating one update
rule as a prerequisite for useful prediction.

\section{Training and Target Diagnostics}
\label{app:training-target-diagnostics}

Predicting a molecular future should require distinguishing the outcome
associated with each input. We examine whether the target geometry makes
these distinctions visible to the cosine objective by removing centering
from one view at a time. We retain $\ell_2$ normalization, predictor dimensions,
loss coefficients, and the four-phase curriculum.
\Cref{tab:centering-all-tasks} reports the complete task results.
The calibration figure uses the 16,949 molecular training targets; the
spectra are computed before rowwise normalization. Average rank in the task
table uses all 17 unrounded metrics with average ranks for ties.

\begin{table*}[t]
\centering
\small
\renewcommand{\arraystretch}{1.1}
\setlength{\tabcolsep}{3pt}
\caption{Target-centering ablation with both predictive views and
$\ell_2$ normalization retained. Editing, ring-system accuracy, and validity are
percentages; property scores are mean improvements; counts are MAE; Murcko
and FTS are similarities. Dark-red bold marks column-best task scores,
including ties at the displayed precision; shading identifies the default
configuration. Lower average rank is better.}
\label{tab:centering-all-tasks}
\resizebox{\textwidth}{!}{%
\begin{tabular}{lrrrrrrrrr}
\toprule
& \multicolumn{3}{c}{Molecule editing (\%) $\uparrow$} & \multicolumn{6}{c}{Property improvement $\uparrow$} \\
\cmidrule(lr){2-4}\cmidrule(lr){5-10}
Target calibration & Add & Delete & Sub. & LogP & Solubility & QED & DRD2 & JNK3 & GSK3-$\beta$ \\
\midrule
\rowcolor{tablemethodhighlight}
\textbf{Both views centered} & \localtie{78.3} & \localwin{86.7} & 61.1 & 0.8318 & 0.9659 & \localwin{0.2296} & \localwin{0.2848} & \localwin{0.1296} & \localwin{0.2515} \\
Without molecular centering & 75.0 & 73.3 & \localwin{62.2} & \localwin{0.8701} & \localwin{0.9703} & 0.2151 & 0.2645 & 0.1163 & 0.2329 \\
Without text centering & \localtie{78.3} & 75.0 & 58.9 & 0.7854 & 0.8311 & 0.2036 & 0.2644 & 0.1174 & 0.2232 \\
\bottomrule
\end{tabular}}
\par\vspace{1.4em}\noindent
\resizebox{\textwidth}{!}{%
\begin{tabular}{lrrrrrrrr}
\toprule
& \multicolumn{4}{c}{Molecular understanding} & \multicolumn{2}{c}{Major product} & \multicolumn{2}{c}{Byproduct} \\
\cmidrule(lr){2-5}\cmidrule(lr){6-7}\cmidrule(lr){8-9}
Target calibration & FG $\downarrow$ & Ring $\downarrow$ & Murcko $\uparrow$ & Ring-sys. (\%) $\uparrow$ & FTS $\uparrow$ & Valid (\%) $\uparrow$ & FTS $\uparrow$ & Valid (\%) $\uparrow$ \\
\midrule
\rowcolor{tablemethodhighlight}
\textbf{Both views centered} & \localtie{0.0667} & \localtie{0.1333} & 0.8065 & 81.11 & \localwin{0.6492} & \localtie{88.33} & \localwin{0.1806} & 45.33 \\
Without molecular centering & \localtie{0.0667} & \localtie{0.1333} & 0.7849 & \localwin{83.89} & 0.6396 & \localtie{88.33} & 0.1773 & 45.00 \\
Without text centering & 0.0700 & 0.1500 & \localwin{0.8115} & 83.33 & 0.6211 & 86.67 & 0.1695 & \localwin{48.33} \\
\bottomrule
\end{tabular}}
\par\vspace{0.8em}\noindent
\begin{tabular}{lr}
\toprule
Target calibration & Avg. rank $\downarrow$ \\
\midrule
Both views centered & \textbf{1.53} \\
Without molecular centering & 1.97 \\
Without text centering & 2.50 \\
\bottomrule
\end{tabular}
\end{table*}

Retaining both centers gives the strongest average rank and outperforms
either removal on most task metrics. Removing text centering also lowers
all six property-improvement scores. This supports the calibrated targets
used in the dual-view model. Centering changes both target geometry and the
numerical scale of the cosine loss, so the comparison evaluates their
combined effect.

\begin{figure*}[t]
    \centering
    \includegraphics[width=\textwidth]{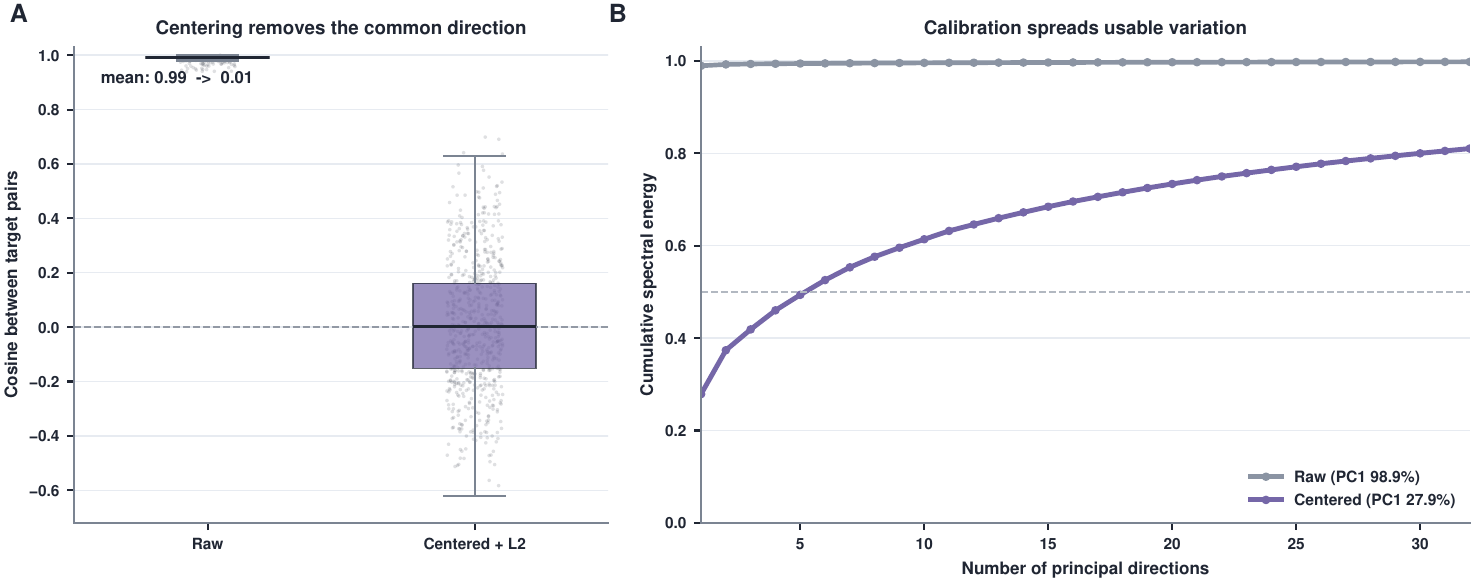}
    \caption{Calibration of molecular prediction targets.
    \textbf{Left:} Pairwise cosine before and after centering and $\ell_2$ normalization.
    \textbf{Right:} Spectral energy of the raw second moment and centered
    covariance, computed before rowwise normalization.
    Appendix~\ref{app:analysis-methods} defines the diagnostics.}
    \label{fig:target-calibration}
\end{figure*}

The raw SMI-TED targets share a dominant direction. Centering followed by
$\ell_2$ normalization reduces mean random-pair cosine from $0.989$ to $0.010$
(\cref{fig:target-calibration}). The spectrum likewise shows that much of the
raw representation energy lies in a common direction. Calibration makes
differences among molecular outcomes more visible to the cosine objective.
The resulting task-level relationships are examined in Appendix~\ref{app:target-structure}.

Autoregressive losses remain similar across Coconut-Chem, text-only,
and dual prediction. At the end of the final curriculum phase, the dual
model's autoregressive, weighted textual, and weighted molecular losses
are approximately $0.0665$, $0.0144$, and $0.0021$. Predictive losses are
lower than their shuffled-source counterparts.

The single-view centering comparisons in \cref{tab:target-centering-ablation}
use coefficients $0.05$ for textual prediction and $0.01$ for molecular
prediction; the dual comparison uses $0.025$ and $0.01$. All four
comparisons use all 17 task metrics. Differences below $10^{-4}$ are
counted as ties, accounting for metric direction. Prediction distances
are averaged over the final 50 training batches.

\begin{table*}[t]
\centering
\small
\renewcommand{\arraystretch}{1.1}
\setlength{\tabcolsep}{7pt}
\caption{Target centering and source-specific prediction.
Task-score changes count improved, unchanged, and worse metrics relative to
the corresponding centered model after removing the indicated center.
The final column gives cosine distances using the
original and shuffled latent sources. Settings and aggregation are described
in the surrounding text.}
\label{tab:target-centering-ablation}
\resizebox{\textwidth}{!}{%
\begin{tabular}{llrrc}
\toprule
Objective & Center removed & Metrics & Task-score changes & Original / shuffled distance \\
\midrule
Text only & Text & 17 & 10 / 3 / 4 & 0.102 / 0.330 \\
Molecular only & Molecular & 17 & 4 / 3 / 10 & 0.003 / 0.009 \\
Dual & Molecular & 17 & 4 / 3 / 10 & 0.003 / 0.009 \\
Dual & Text & 17 & 3 / 1 / 13 & 0.095 / 0.217 \\
\bottomrule
\end{tabular}}
\end{table*}

Without molecular centering, original- and shuffled-source distances are
both small ($0.003$ and $0.009$), consistent with predicting a shared target
direction rather than relying strongly on the corresponding input.
Text targets retain more source specificity, although its measured separation
also decreases without centering. Removing text centering helps some text-only
task scores, whereas the dual model benefits on more metrics from retaining
both centers.
Together, these diagnostics suggest that a target's usefulness depends on
both its encoded information and the outcome differences emphasized by the
prediction loss. This motivates considering the encoder, calibration, and
discrepancy together when designing scientific prediction targets
(Appendix~\ref{app:additional-related}).

\section{Prediction Views and Chemical Task Performance}
\label{app:prediction-view-ablation}

The two predictive views describe a solution in different spaces. To examine
what each contributes, we compare textual prediction, molecular prediction,
and their combination in Coconut-Chem
(\cref{tab:prediction-view-performance}). All configurations
use six latent thoughts, three epochs per curriculum stage, and the ChemCoTBench evaluation specified in Appendix~\ref{app:integrity}.
Decoding uses temperature $0.7$, top-$p=0.9$, three responses per example. The molecular coefficient is $0.01$ in both
relevant variants, while the textual coefficient is $0.05$ for text-only
and $0.025$ for dual prediction. This comparison therefore varies textual
supervision strength as well as the set of views. Its average rank uses
the 17 metrics available for all three models,
from each, with unrounded scores and average ranks for ties.

\begin{table*}[t]
\centering
\small
\setlength{\tabcolsep}{3pt}
\renewcommand{\arraystretch}{1.1}
\caption{Chemical performance with different prediction views.
Editing, ring-system accuracy, and validity are percentages; property scores
are mean improvements; counts are MAE.
Dark-red bold marks column-best task scores, including ties at the displayed
precision; shading identifies dual prediction. Loss coefficients and rank
aggregation are specified in the text.}
\label{tab:prediction-view-performance}
\resizebox{\textwidth}{!}{%
\begin{tabular}{lrrrrrrrrr}
\toprule
Prediction views & Add $\uparrow$ & Delete $\uparrow$ & Sub. $\uparrow$ & LogP $\uparrow$ & Solubility $\uparrow$ & QED $\uparrow$ & DRD2 $\uparrow$ & JNK3 $\uparrow$ & GSK3-$\beta$ $\uparrow$ \\
\midrule
Textual & 81.7 & 80.0 & 58.9 & 0.7839 & 0.9178 & 0.2229 & 0.2684 & 0.1100 & 0.2415 \\
Molecular (SMI-TED) & \localwin{83.3} & \localwin{90.0} & \localwin{62.8} & \localwin{0.8624} & 0.9359 & \localwin{0.2333} & 0.2579 & 0.1141 & 0.2442 \\
\rowcolor{tablemethodhighlight}
Textual + molecular & 78.3 & 86.7 & 61.1 & 0.8318 & \localwin{0.9659} & 0.2296 & \localwin{0.2848} & \localwin{0.1296} & \localwin{0.2515} \\
\bottomrule
\end{tabular}}
\par\vspace{1.4em}\noindent
\resizebox{\textwidth}{!}{%
\begin{tabular}{lrrrrrrrr}
\toprule
Prediction views & FG $\downarrow$ & Ring $\downarrow$ & Murcko $\uparrow$ & Ring-sys. $\uparrow$ & Major FTS $\uparrow$ & Major valid $\uparrow$ & Byproduct FTS $\uparrow$ & Byproduct valid $\uparrow$ \\
\midrule
Textual & 0.0867 & 0.1500 & 0.7856 & \localtie{81.67} & 0.6036 & 83.67 & 0.1445 & 44.00 \\
Molecular (SMI-TED) & 0.0767 & 0.1500 & 0.7552 & \localtie{81.67} & 0.6392 & 86.67 & 0.1741 & 45.00 \\
\rowcolor{tablemethodhighlight}
Textual + molecular & \localwin{0.0667} & \localwin{0.1333} & \localwin{0.8065} & 81.11 & \localwin{0.6492} & \localwin{88.33} & \localwin{0.1806} & \localwin{45.33} \\
\bottomrule
\end{tabular}}
\par\vspace{0.8em}\noindent
\begin{tabular}{lr}
\toprule
Prediction views & Avg. rank $\downarrow$ \\
\midrule
Textual & 2.71 \\
Molecular (SMI-TED) & 1.82 \\
Textual + molecular & \textbf{1.47} \\
\bottomrule
\end{tabular}
\end{table*}

Molecular prediction is particularly effective for editing, while dual
prediction gives the strongest average rank over the metrics shared by
all three variants. Adding the textual view improves most property
objectives and all reported reaction metrics relative to molecular-only
prediction. The combined supervision thus benefits a broader set of tasks,
while the full table retains the cases favoring a single view. These
patterns suggest that the usefulness of a future view depends on the
information needed to solve the task.

\paragraph{Transfer to Vanilla Coconut.}
\label{app:vanilla-transfer}
Vanilla Coconut receives source SMILES through the textual prompt without
source ChemTokens. The four variants in
\cref{tab:vanilla-prediction-view-performance} share a text-only initializer,
six latent thoughts, three epochs per curriculum phase,
and the 1,120-example evaluation at temperature $0.7$, top-$p=0.9$, with
three responses per example. Average rank uses all 17 unrounded metrics and
average ranks for ties.

\begin{table*}[t]
\centering
\small
\setlength{\tabcolsep}{3pt}
\renewcommand{\arraystretch}{1.1}
\caption{Prediction-view comparison in Vanilla Coconut without
source ChemTokens. Editing, ring-system accuracy, and validity are percentages;
property scores are mean improvements; counts are MAE. Training and evaluation
settings are given in the text. Dark-red bold marks column-best task scores,
including ties at the displayed precision; shading identifies textual
prediction. Lower average rank is better.}
\label{tab:vanilla-prediction-view-performance}
\resizebox{\textwidth}{!}{%
\begin{tabular}{lrrrrrrrrr}
\toprule
& \multicolumn{3}{c}{Molecule editing (\%) $\uparrow$} & \multicolumn{6}{c}{Property improvement $\uparrow$} \\
\cmidrule(lr){2-4}\cmidrule(lr){5-10}
Prediction views & Add & Delete & Sub. & LogP & Solubility & QED & DRD2 & JNK3 & GSK3-$\beta$ \\
\midrule
Vanilla Coconut & 71.7 & 88.3 & 56.1 & 0.7726 & 0.7250 & 0.1968 & 0.2300 & 0.0861 & 0.1861 \\
\rowcolor{tablemethodhighlight}
Textual & \localwin{76.7} & 86.7 & 56.1 & \localwin{0.8656} & \localwin{0.9001} & 0.2048 & 0.2268 & \localwin{0.0992} & 0.1875 \\
Molecular (SMI-TED) & 68.3 & 85.0 & \localwin{57.2} & 0.7960 & 0.8107 & 0.2105 & \localwin{0.2420} & 0.0769 & \localwin{0.2153} \\
Textual + molecular & 66.7 & \localwin{90.0} & 55.6 & 0.8429 & 0.8799 & \localwin{0.2307} & 0.2376 & 0.0796 & 0.1972 \\
\bottomrule
\end{tabular}}
\par\vspace{1.4em}\noindent
\resizebox{\textwidth}{!}{%
\begin{tabular}{lrrrrrrrr}
\toprule
& \multicolumn{4}{c}{Molecular understanding} & \multicolumn{2}{c}{Major product} & \multicolumn{2}{c}{Byproduct} \\
\cmidrule(lr){2-5}\cmidrule(lr){6-7}\cmidrule(lr){8-9}
Prediction views & FG $\downarrow$ & Ring $\downarrow$ & Murcko $\uparrow$ & Ring-sys. $\uparrow$ & FTS $\uparrow$ & Valid $\uparrow$ & FTS $\uparrow$ & Valid $\uparrow$ \\
\midrule
Vanilla Coconut & \localtie{0.0600} & \localtie{0.1000} & 0.7537 & \localwin{81.67} & 0.5673 & 82.67 & 0.1648 & 47.00 \\
\rowcolor{tablemethodhighlight}
Textual & \localtie{0.0600} & \localtie{0.1000} & 0.7485 & 77.22 & \localwin{0.6106} & \localwin{86.33} & 0.1849 & \localwin{50.00} \\
Molecular (SMI-TED) & 0.0700 & \localtie{0.1000} & 0.7045 & 77.78 & 0.5809 & 81.33 & 0.1596 & 46.33 \\
Textual + molecular & 0.0667 & \localtie{0.1000} & \localwin{0.7848} & 79.44 & 0.5441 & 78.00 & \localwin{0.1893} & 47.00 \\
\bottomrule
\end{tabular}}
\par\vspace{0.8em}\noindent
\begin{tabular}{lr}
\toprule
Prediction views & Avg. rank $\downarrow$ \\
\midrule
Vanilla Coconut & 2.65 \\
Textual & \textbf{2.09} \\
Molecular (SMI-TED) & 2.85 \\
Textual + molecular & 2.41 \\
\bottomrule
\end{tabular}
\end{table*}

Textual prediction improves 10 of 17 metrics over Vanilla Coconut, with
four regressions and three ties; molecular and dual prediction
improve seven and eight metrics, respectively. Textual prediction has the lowest average rank in this comparison,
while molecular and dual prediction favor different tasks. For example,
molecular prediction performs best on DRD2 and GSK3-$\beta$, while dual
prediction performs best on QED and Murcko similarity. These results show
that predictive supervision can benefit a model whose molecular inputs are
provided only as text. The different patterns in Coconut-Chem and Vanilla
Coconut motivate choosing future views together with the task and reasoning
setup; their source interfaces, initialization, and training histories are
not otherwise matched. Appendix~\ref{app:terminal-organization} examines the associated
representation patterns within each setting.

\clearpage
\section{Chemical Relationships in the Prediction Targets}
\label{app:target-structure}

\begin{figure*}[t]
    \centering
    \includegraphics[width=0.92\textwidth]{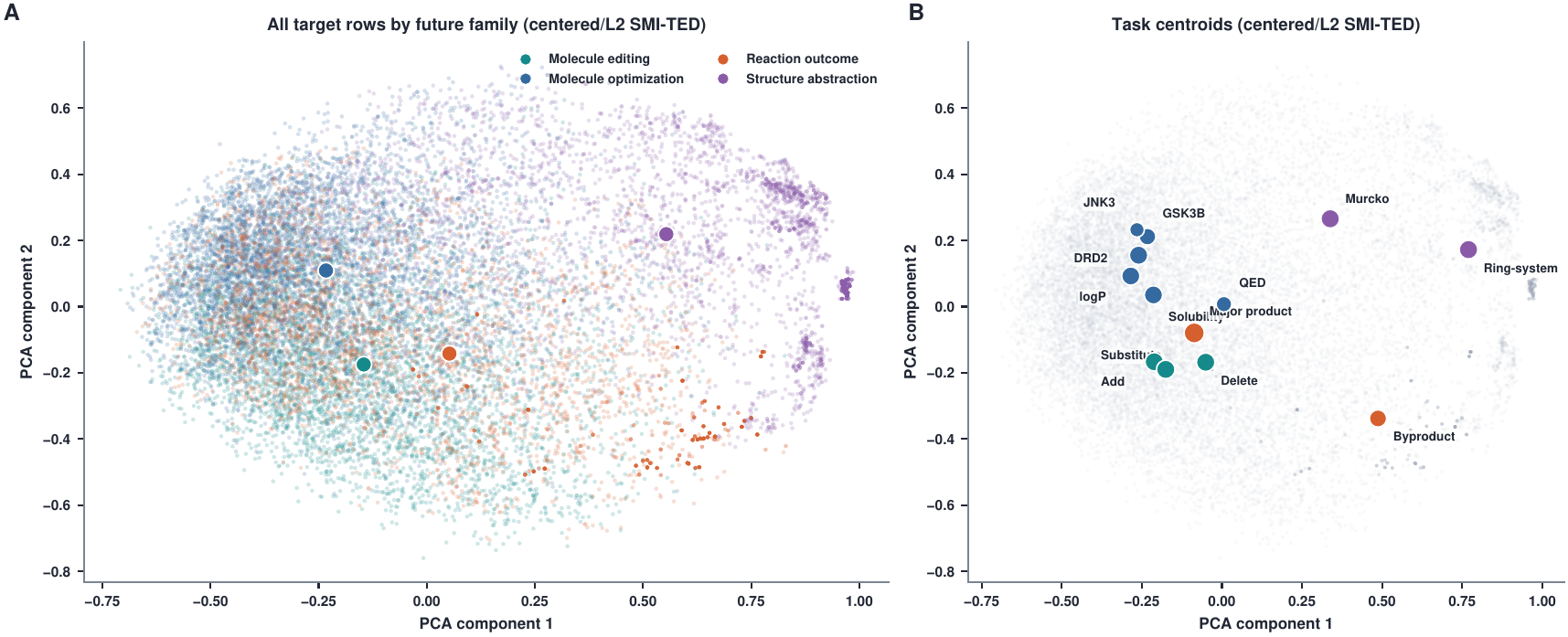}
    \caption{Structure of the calibrated molecular target space.
    \textbf{A:} Target PCA colored by task family. \textbf{B:} Task centroids
    in the same projection. Both panels describe the target encoder's
    representations, not the reasoner's latent states.}
    \label{fig:target-task-structure}
\end{figure*}

\begin{figure*}[t]
    \centering
    \includegraphics[width=0.92\textwidth]{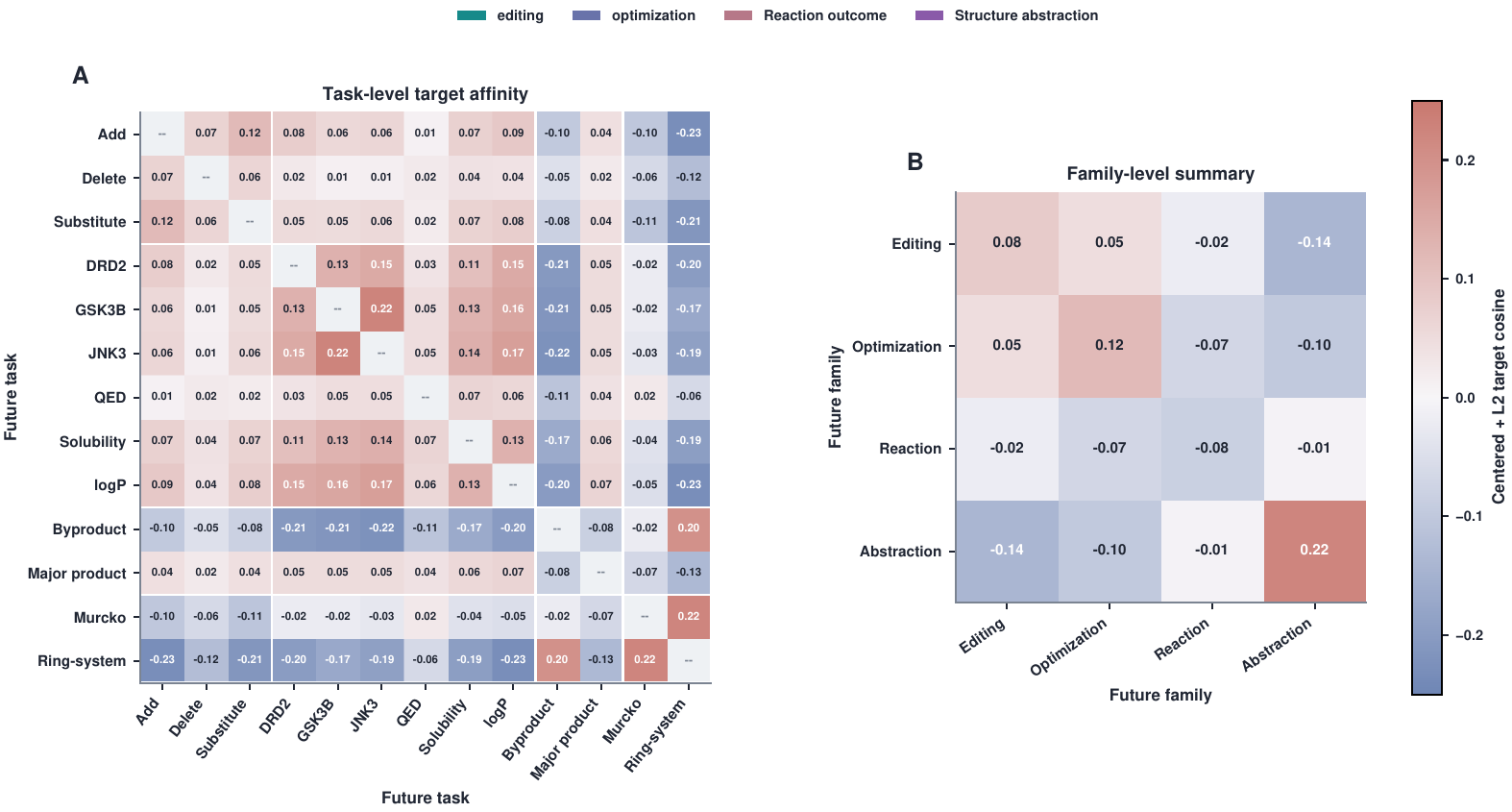}
    \caption{Molecular target relationships across tasks.
    \textbf{Left:} Mean cross-task cosine after centering and $\ell_2$ normalization;
    the within-task diagonal is omitted. \textbf{Right:} Aggregation by task
    family. Definitions are given in Appendix~\ref{app:geometry-statistics}.}
    \label{fig:target-affinity}
\end{figure*}

To understand what the molecular encoder supplies as supervision, we examine
the relationships among its representations of reference outcomes. We use the
calibrated SMI-TED embeddings of 16,949 molecular training targets, before
considering the latent states trained to predict them. The shared PCA in
\cref{fig:target-task-structure} shows individual targets and task centroids;
\cref{fig:target-affinity} summarizes their cross-task cosine relations.
Appendix~\ref{app:geometry-statistics} defines the affinity and task-family aggregation.

The target relations do not simply reproduce the task taxonomy. Murcko and
ring-system targets form a close pair, and several optimization tasks have
positive affinities. Yet major-product and byproduct targets, which share a
reaction-task label, have lower affinity than byproduct and ring-system
targets. The encoder thus relates outcomes in ways that differ from grouping
them by the requested operation.

These relations help explain the role of the molecular target: it supplies
similarities among solutions beyond their task membership. Predicting this view may encourage latent thoughts to reflect
such relationships while distinguishing individual outcomes. We examine
these possibilities in \cref{app:terminal-organization}. The affinities
describe this encoder and target cohort; identifying their specific chemical
drivers would require additional molecular annotations.
\section{Organization of Latent Thoughts}
\label{app:terminal-organization}

After characterizing the targets, we ask how the choice of predictive view
relates to the outcome information in the final latent thought. We compare
Coconut-Chem with textual, molecular, and dual prediction on 16,943 training
examples across 13 tasks. 
We analyze final-layer hidden states at the sixth latent thought.
The PCA is shared across the four variants, while dispersion and task-centroid
agreement use full-dimensional calibrated states. Retrieval uses ridge probes
with 3,390 examples from this training cohort excluded from probe fitting,
and distinct future-SMILES candidates within each task. This is a separate
probe split from \cref{fig:heldout-decodability}; preprocessing and metric
definitions are specified in Appendix~\ref{app:analysis-methods}.

\begin{figure*}[t]
    \centering
    \includegraphics[width=\textwidth]{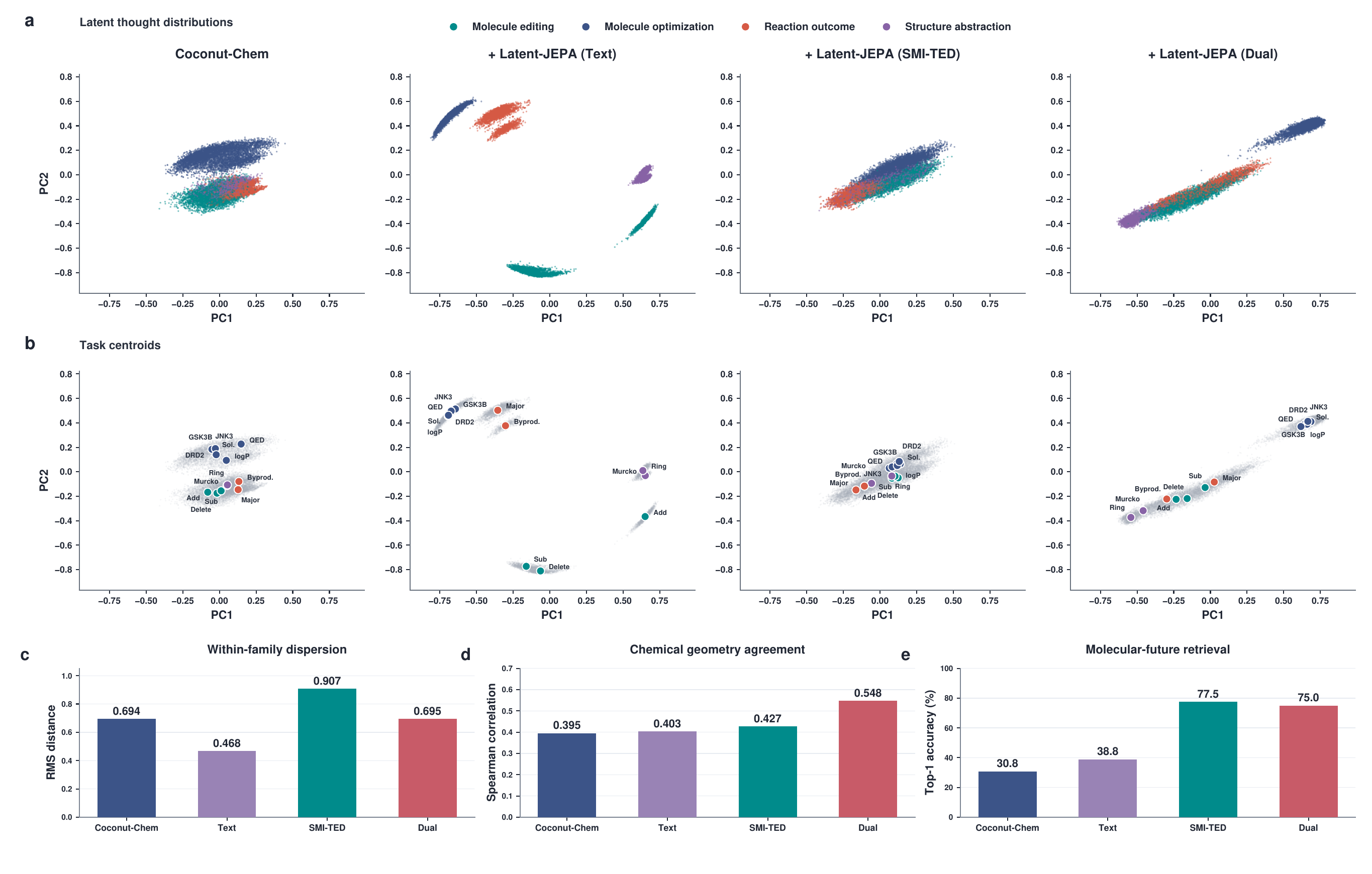}
    \caption{Latent organization under different predictive views.
    \textbf{a--b:} Shared PCA and task centroids for Coconut-Chem and its
    predictive variants. \textbf{c:} Within-family dispersion.
    \textbf{d:} Agreement with molecular task-centroid relations.
    \textbf{e:} Within-task molecular-outcome retrieval on a probe holdout from
    the training cohort. Dispersion and centroid agreement use full-dimensional
    calibrated states; settings are given in the surrounding text.}
    \label{fig:terminal-latent-organization}
\end{figure*}

Molecular prediction yields the highest outcome retrieval, while dual
prediction has the strongest agreement with molecular task-centroid relations
among these configurations (\cref{fig:terminal-latent-organization}). Dual
prediction also raises retrieval from $30.8\%$ to $75.0\%$ relative to
Coconut-Chem. The dispersion statistic describes a different pattern: textual
prediction has the lowest within-family dispersion, and the dual model is
close to Coconut-Chem ($0.695$ versus $0.694$). Thus, the ordering by
dispersion does not follow the ordering by retrieval in this comparison.

\begin{figure*}[t]
    \centering
    \includegraphics[width=\textwidth]{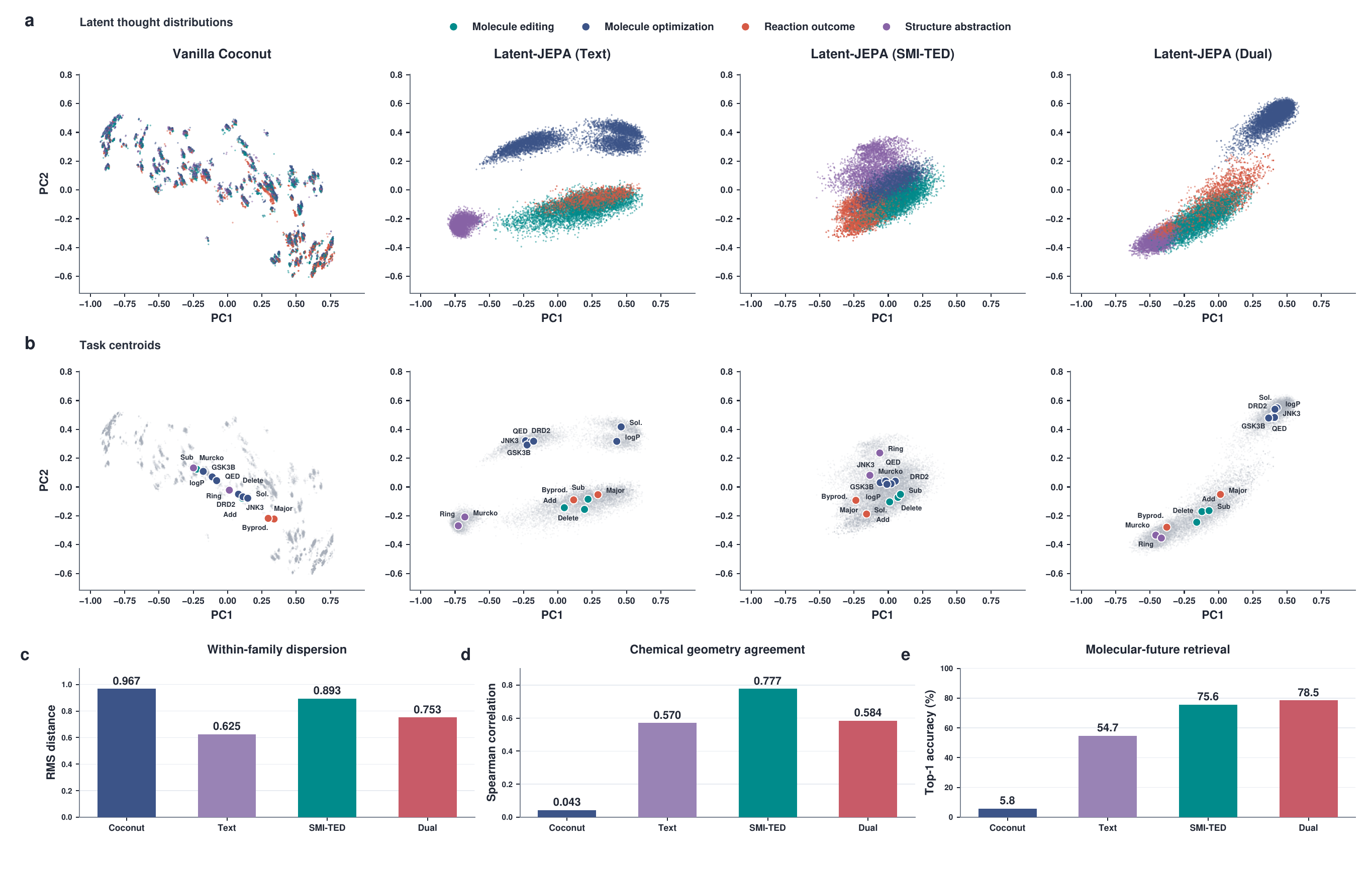}
    \caption{Latent organization in Vanilla Coconut, which receives
    SMILES as text without source ChemTokens. Panels follow
    \cref{fig:terminal-latent-organization}. The PCA basis is shared within this
    comparison, not across architectures. Settings and the probe protocol are
    specified in the text and Appendix~\ref{app:analysis-methods}.}
    \label{fig:vanilla-terminal-latent-organization}
\end{figure*}

We repeat the analysis in Vanilla Coconut to examine target-side supervision
when source molecules enter only as SMILES in the prompt. Dual prediction gives the
highest retrieval, increasing it from $5.8\%$ to $78.5\%$, while molecular
prediction gives the strongest agreement with molecular task relations.
Textual prediction again has the lowest within-family dispersion. The
within-setting comparison shows that outcome information can become more
accessible without a dedicated source-molecule interface. It therefore
supports studying domain encoders as sources of predictive supervision as
well as input representations (Appendix~\ref{app:additional-related}). Coconut-Chem
and Vanilla Coconut also differ in initialization and training history, so
their absolute scores do not isolate the role of ChemTokens.

\begin{figure*}[t]
    \centering
    \includegraphics[width=\textwidth]{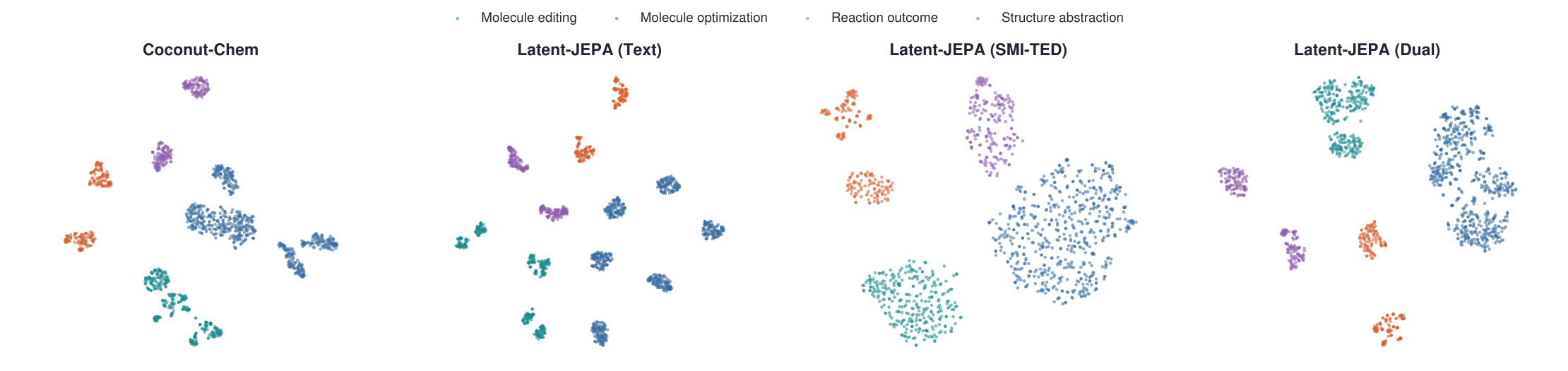}
    \vspace{0.4em}
    \includegraphics[width=\textwidth]{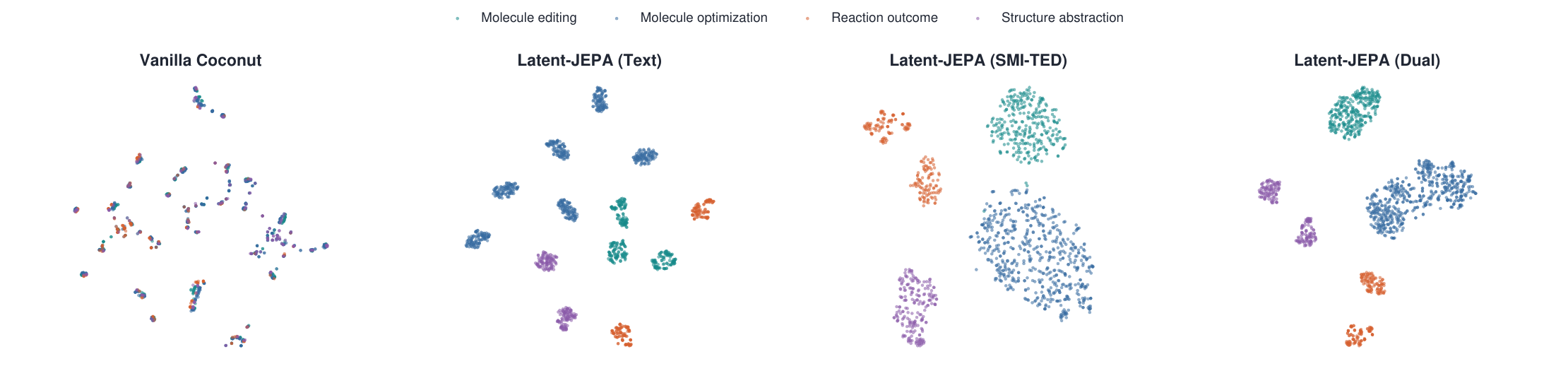}
    \caption{Exploratory local neighborhoods of latent thoughts in
    Coconut-Chem (top) and Vanilla Coconut (bottom). \Cref{fig:terminal-latent-organization,fig:vanilla-terminal-latent-organization}
    provide the full-dimensional quantitative comparisons; preprocessing is
    given in Appendix~\ref{app:analysis-methods}.}
    \label{fig:terminal-latent-tsne}
\end{figure*}

\Cref{fig:terminal-latent-tsne} provides a qualitative view of how predictive supervision changes
local neighborhood organization. Textual prediction produces clearer
task-family separation, while molecular and dual supervision yield different
local arrangements that accompany their stronger molecular-outcome retrieval
and geometry agreement.

\section{Predicting Molecular Outcomes from Latent Thoughts}
\label{app:predictive-diagnostics}
The average probe gains leave open whether the improvement is widespread or concentrated in a few examples. We examine this distribution, then study how prediction and structural agreement vary across tasks and recurrent depth. For probing, we fit a separate ridge map at each step using 1,224 training examples and evaluate it on 321 molecular-target-eligible ChemCoTBench test examples. We compare corresponding targets with a fixed within-task permutation of evaluation targets. The probes use latent hidden states, whereas RSA uses recurrent feedback embeddings. SMI-TED RSA uses the same 321 test examples, and Morgan RSA uses the subset of 294 with valid outcome fingerprints; Appendix~\ref{app:analysis-methods} specifies the protocols and depth indexing.

\begin{figure*}[t]
    \centering
    \includegraphics[width=0.62\textwidth]{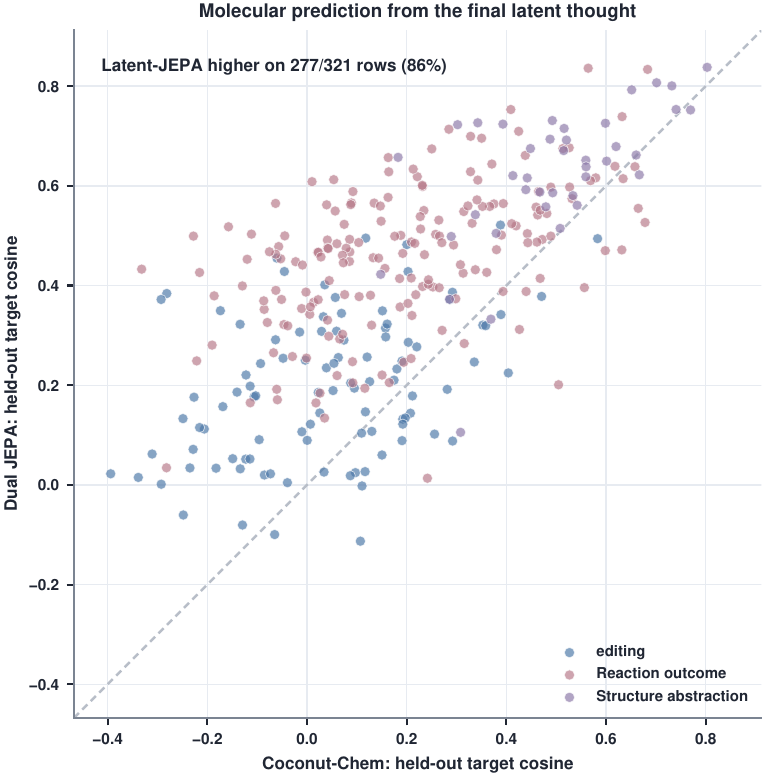}
    \caption{Example-wise molecular-outcome prediction from the final
    latent thought. Each point compares probe cosine for Coconut-Chem
    (horizontal) and dual-view \method{} (vertical); colors denote task families.
    Points above the diagonal favor \method{}. The probe split is shared with
    \cref{fig:heldout-decodability}.}
    \label{fig:terminal-readout-scatter}
\end{figure*}

The improvement extends across most of the probe-evaluation cohort:
\method{} increases cosine on 277 of 321 examples
(\cref{fig:terminal-readout-scatter}), with mean final-update cosine
rising from $0.186$ to $0.397$. Corresponding targets remain more predictable
than targets reassigned within the same task. The higher mean therefore
reflects improved outcome prediction across many examples rather than a
small subset.

\begin{figure*}[t]
    \centering
    \includegraphics[width=0.92\textwidth]{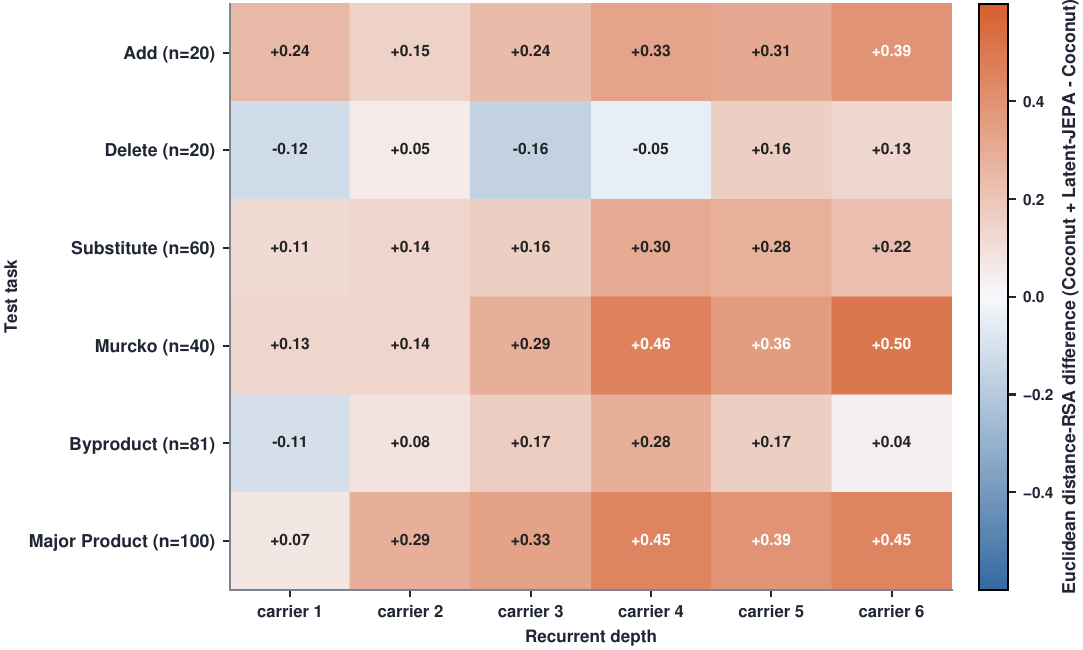}
    \caption{Task-wise changes in molecular geometry across recurrent
    depth. Each cell gives dual-view \method{} minus Coconut-Chem within-task
    RSA, using feedback-embedding and calibrated molecular-target distances.
    Positive values indicate greater agreement for \method{}; these are
    between-model differences, not absolute RSA trajectories.}
    \label{fig:chemical-rsa-heatmap}
\end{figure*}

We next examine whether the aggregate geometry trend extends across tasks.
\Cref{fig:chemical-rsa-heatmap} shows positive final-update gains over
Coconut-Chem in every analyzed task. The relative gains are larger at later
updates for Murcko and major-product prediction, while deletion and
byproduct prediction follow different depth patterns.

At the aggregate level, dual-model SMI-TED RSA rises from $0.190$ at the
first update to $0.400$ at the final update. Final-update RSA is $0.110$
for Coconut-Chem, $0.115$ for text-only prediction, and $0.392$ for
molecular-only prediction. With Morgan fingerprints, the final-update scores
are $0.029$, $0.118$, and $0.329$ for Coconut-Chem, text-only, and dual
prediction. The task breakdown therefore adds detail to the overall
structural agreement measured in \cref{fig:chemical-geometry}.

\begin{figure*}[t]
    \centering
    \includegraphics[width=0.92\textwidth]{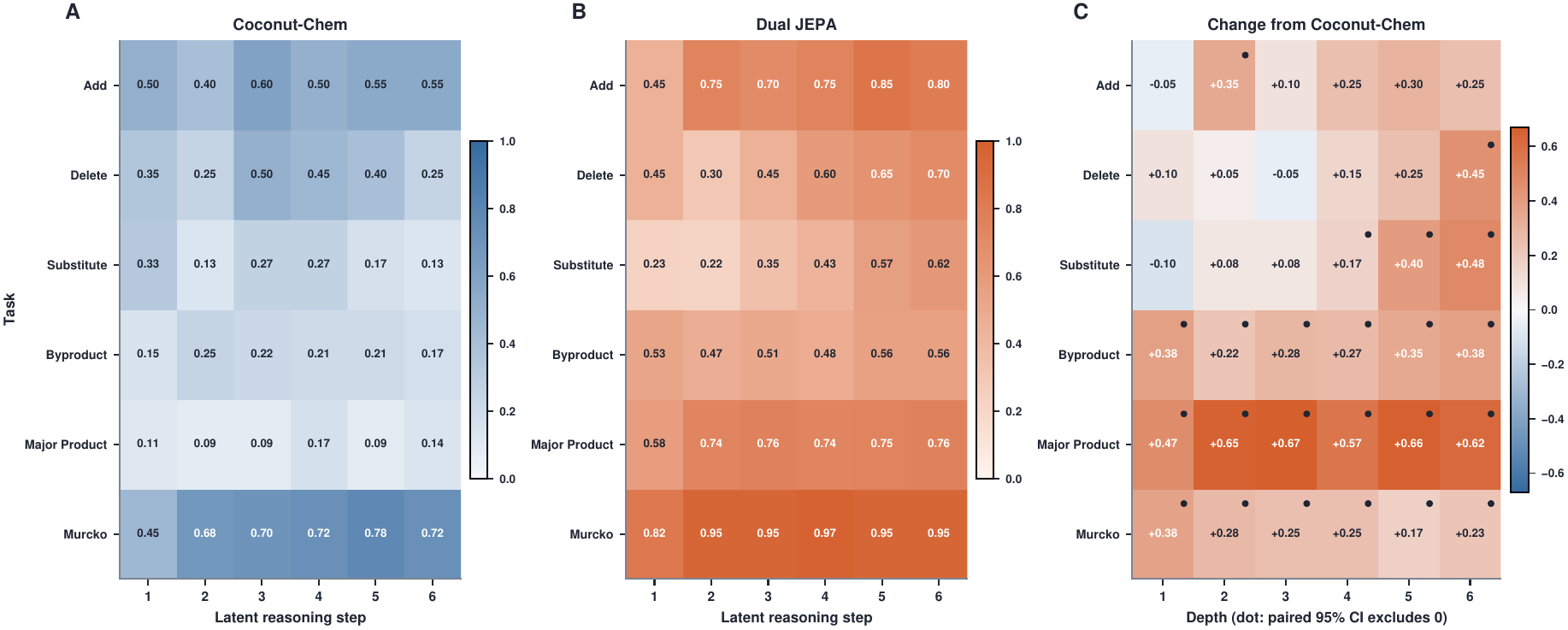}
\caption{Molecular-outcome retrieval by task and recurrent depth.
    \textbf{A:} Coconut-Chem Hit@5. \textbf{B:} Dual-view \method{} Hit@5.
    \textbf{C:} Their difference; dots mark 95\% bootstrap intervals excluding
    zero. The heatmaps use the same latent hidden states, probe holdout, and
    within-task candidates as \cref{fig:heldout-decodability}. Averaging each
    column over tasks recovers the corresponding aggregate Hit@5 curve;
    Appendix~\ref{app:analysis-methods} gives the protocol.}
    \label{fig:task-depth-decodability}
\end{figure*}

\Cref{fig:task-depth-decodability} shows that retrieval gains occur across
tasks and are not confined to the final update. In the aggregate analysis,
the dual model's probe cosine and Hit@5 generally increase across recurrent
steps, with smaller changes near the final updates
(\cref{fig:heldout-decodability}). Its structural agreement also increases
overall (\cref{fig:chemical-geometry}). The task-level retrieval gains follow
different depth profiles: gains for substitution become larger at later
steps, whereas major-product and Murcko-scaffold prediction already show
clear gains at the first step. These task-dependent patterns motivate
studying predictive objectives whose horizons or levels of abstraction
vary across recurrent depth.

\clearpage
\section{Explicit Reasoning after Latent Computation}
\label{app:residual-reasoning}

\begin{figure*}[t]
    \centering
    \includegraphics[width=0.78\textwidth]{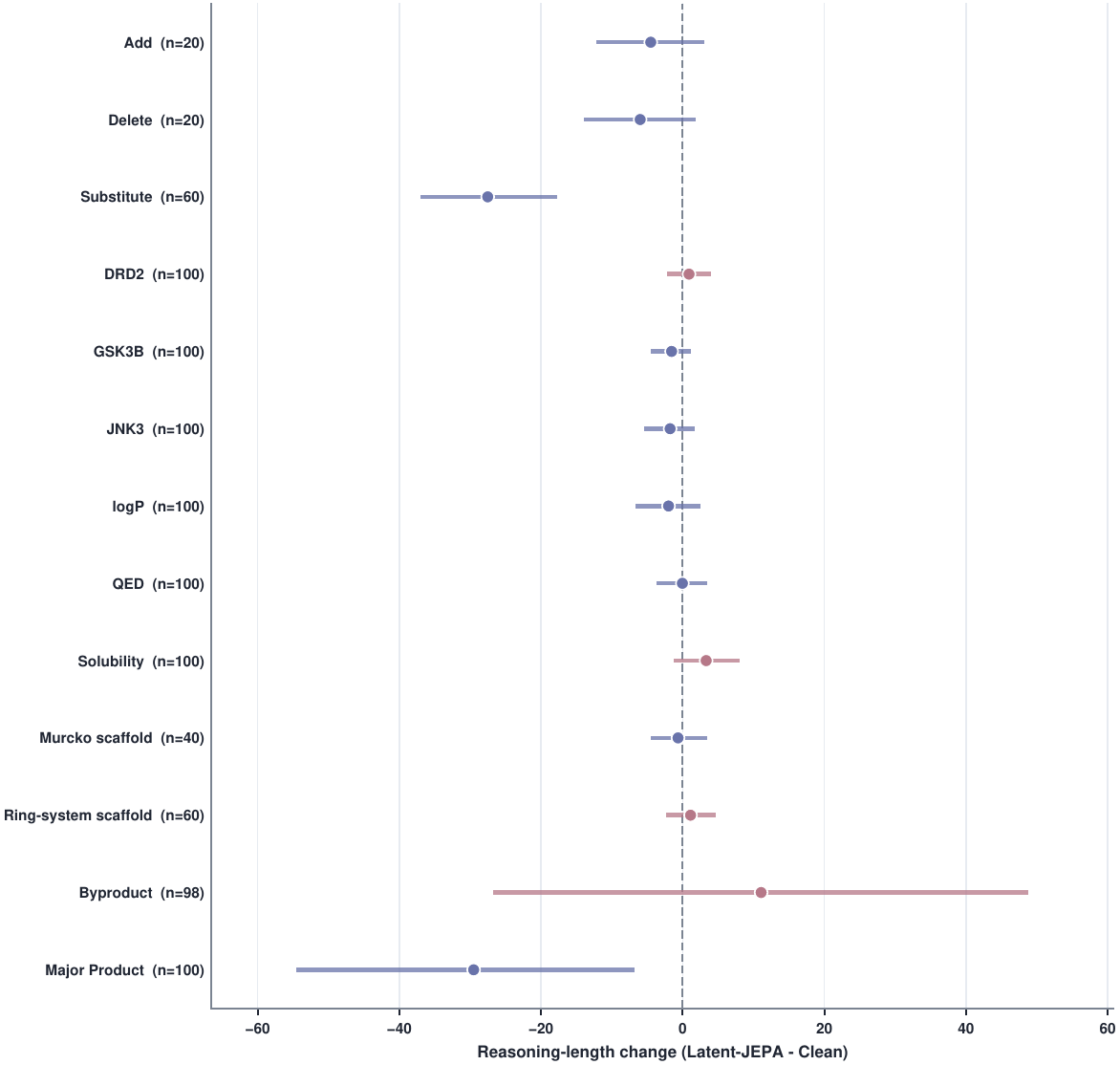}
    \caption{Change in explicit reasoning length by task. Points show
    the difference in mean pre-answer token count between \method{} and
    Coconut-Chem, with bootstrap intervals. Negative values indicate shorter
    visible reasoning; token exclusions are specified in the text.}
    \label{fig:residual-reasoning}
\end{figure*}

The preceding analyses concern latent states before answer generation.
We also examine whether predictive supervision changes how much reasoning
the model subsequently expresses in text. We compare the mean number of
pre-answer tokens, excluding final SMILES and answer-declaration lines.
\Cref{fig:residual-reasoning} reports the task-wise difference between
\method{} and Coconut-Chem with bootstrap intervals.

The clearest reductions occur in substitution and major-product prediction,
whereas optimization tasks remain close to the baseline and byproduct
prediction has a wider interval. These changes provide a behavioral
counterpart to the representation analyses: predictive supervision can alter
how a solution is expressed, with no uniform reduction across tasks.

\clearpage
\section{Methods for Representation Analysis}
\label{app:analysis-methods}

\phantomsection\label{app:organization-settings}
This appendix specifies the representations, cohorts, and measurements used
to analyze molecular-outcome recoverability, molecular relationships, and
task organization. Probing and RSA use the matched-view Coconut-Chem
configurations described in Appendix~\ref{app:terminal-organization};
optimization and curriculum settings are given in Appendix~\ref{app:contract}.

\subsection{Representations and evaluation cohorts}
\label{app:representation-cohorts}

We distinguish the final-layer hidden state at a continuous-thought position
from the feedback embedding supplied to the next recurrent computation.
The former is the supervised readout and is used for linear probing; the
latter is used for the RSA curves. On our fixed-length evaluation route,
the hidden state at step $t$ supplies the feedback embedding at step $t+1$.
Neither view receives the gold continuation during evaluation extraction.

\Cref{tab:analysis-cohorts} distinguishes training data used to fit probes
or describe task organization from test data used to evaluate molecular
outcomes. Molecular targets use
the training-mean calibration in \cref{eq:chem-target}.

\begin{table}[t]
\centering
\small
\caption{Cohorts used for representation analysis, with counts in examples.}
\label{tab:analysis-cohorts}
\begin{tabular}{lp{0.53\textwidth}}
\toprule
Analysis & Cohort \\
\midrule
Main linear probing and Hit@5 & 1,224 training examples across 13 tasks for probe fitting; 321 test examples across six tasks for evaluation \\
SMI-TED RSA & 321 molecular-target-eligible test examples across six tasks \\
Morgan RSA & 294 test examples with valid molecular outcomes across the same six tasks \\
Latent PCA and task organization & 16,943 training examples; the separate retrieval analysis holds out 3,390 from probe fitting \\
Target calibration and task affinity & 16,949 molecular training targets \\
\bottomrule
\end{tabular}
\end{table}

\subsection{Linear probing and molecular retrieval}
\label{app:probe-methods}

For each model and reasoning step, we fit a separate ridge probe:

\begin{equation}
 (A,b)=\arg\min_{A,b}\sum_{i\in\mathcal D_{\mathrm{fit}}}
 \|A u_i+b-z_i\|_2^2+\alpha\|A\|_F^2,
 \qquad \widehat z_i=\frac{A u_i+b}{\|A u_i+b\|_2},
 \label{eq:analysis-probe}
\end{equation}
where $z_i$ is the normalized molecular target. For the main probe figure,
$u_i$ is a 512-dimensional sparse random projection of the latent hidden
state after featurewise standardization. Means and standard deviations are
estimated only on the probe-fitting split. Models and steps share the random
projection and fitting examples, but not fitted probe weights. We use ridge
coefficient $\alpha=10$, an unpenalized intercept, and the LSQR solver with tolerance
$10^{-4}$. Training examples are selected by task-stratified hashing of
example keys. Test examples do not enter probe fitting or selection of probe
hyperparameters. We report example-mean cosine $\widehat z_i^\top z_i$.
These probes are fitted after model training, independently of the JEPA
predictors.

For retrieval, let $\mathcal C_q$ contain one target per distinct stored
future-SMILES string among evaluation examples of task $q$. Each evaluation
example's target belongs to its candidate set. Candidates receive cosine
scores $s_{ij}=\widehat z_i^\top z_j$. The reported rank and Hit@$k$ are:

\begin{equation}
 r_i=1+\sum_{j\in\mathcal C_{q(i)}}
 \mathbf 1[s_{ij}>s_{i,i}+10^{-8}],\qquad
 \mathrm{Hit@}k=\frac1{|\mathcal Q|}\sum_{q\in\mathcal Q}
 \frac1{n_q}\sum_{i:q(i)=q}\mathbf 1[r_i\leq k].
 \label{eq:analysis-hit}
\end{equation}
Here $s_{i,i}$ denotes the score of the example's corresponding candidate,
including when several examples share that candidate. Scores tied within
the numerical tolerance do not count as outranking it. Main-figure candidate
sets have 20, 20, 60, 40, 100, and 61 entries for addition, deletion,
substitution, Murcko-scaffold, major-product, and byproduct tasks,
respectively. Uniform random ranking gives the task-average
baseline $|\mathcal Q|^{-1}\sum_q\min(k,|\mathcal C_q|)/|\mathcal C_q|$,
which is $0.1401$ for $k=5$. The task-organization figures instead report
example-averaged Top-1 retrieval on their separate probe holdout
from the training cohort, with candidates drawn from that holdout.

For the shuffled-target panel, we hold the fitted probes and their predictions
fixed and randomly reassign evaluation targets within each task.
Both models use the same permutation. This preserves task membership
while disrupting example-level correspondence; the ordinary permutation
retains self-matches and repeated molecular outcomes.

\subsection{Representational similarity analysis}
\label{app:rsa-methods}

RSA measures whether examples that are close in one representation space are
also close in another.
For task $q$ and reasoning step $t$, form two vectors over all distinct
unordered example pairs:

\begin{equation}
 a_{q,t}=\bigl(d_h(h_{i,t},h_{j,t})\bigr)_{i<j,\ i,j\in q},\qquad
 b_q=\bigl(d_z(z_i,z_j)\bigr)_{i<j,\ i,j\in q}.
\end{equation}
We calculate

\begin{equation}
 \mathrm{RSA}_{q,t}=\mathrm{corr}\bigl(\mathrm{rank}(a_{q,t}),
 \mathrm{rank}(b_q)\bigr),\qquad
 \mathrm{RSA}_{t}=\frac1{|\mathcal Q|}\sum_{q\in\mathcal Q}
 \mathrm{RSA}_{q,t},
 \label{eq:analysis-rsa}
\end{equation}
using Spearman rank correlation with average ranks for ties. Higher RSA
indicates stronger agreement in relative distances. We first compute RSA
within each task, then average equally across tasks. The task-depth
heatmap reports the difference between \method{} and Coconut-Chem RSA in each cell.

The SMI-TED analysis uses Euclidean distances between feedback embeddings
and between training-mean-centered, L2-normalized molecular targets. The
Morgan analysis instead uses latent cosine distance and fingerprint Tanimoto
distance, $1-|f_i\cap f_j|/|f_i\cup f_j|$, for bit fingerprints of radius two
and length 2,048 with chirality enabled.
Morgan fingerprints encode local atom neighborhoods; $f_i$ denotes the set
of active bit positions for molecule $i$. Tanimoto similarity is the
fraction of active bits shared by the two fingerprints relative to their
union, ranging from zero (no shared bits) to one (identical fingerprints).
The distance is one minus this similarity. Identical fingerprints do not
necessarily imply identical molecules.

Using cosine distance for the SMI-TED comparison preserves the ordering
between models. First- and final-update RSA are $0.176$ and $0.410$ for
\method{}, compared with $0.138$ and $0.118$ for Coconut-Chem.

\subsection{Projections and geometry statistics}
\label{app:geometry-statistics}

For task organization, we use final-layer hidden states at the sixth latent
thought, subtract each model's cohort mean, and L2-normalize each row.
We fit one PCA basis across the four variants within each architecture.
Task centroids are means of projected samples, with common axis limits
within a comparison. Dispersion and task-centroid agreement use the
full-dimensional calibrated states.

For t-SNE, we sample up to 100 examples per task, reduce to 50 PCA dimensions,
and fit each model's projection with perplexity 30.

For a family $g$ of normalized states $u_i$, its RMS dispersion is
$\sqrt{|g|^{-1}\sum_{i\in g}\|u_i-\bar u_g\|_2^2}$, where $\bar u_g$ is
the family mean. We average this statistic equally over four families.
Smaller values indicate less variation around the family mean.
Task-centroid agreement correlates the cosine
relations among the 13 latent task means with those among molecular task
means, using the 78 unordered task pairs. Target affinity between two
distinct tasks is the mean cosine over all cross-task target pairs.

For a positive-semidefinite matrix with eigenvalues $\lambda_j$, let
$p_j=\lambda_j/\sum_l\lambda_l$. Its spectral effective rank is

\begin{equation}
 r_{\mathrm{eff}}=\exp\!\left(-\sum_{j:p_j>0}p_j\log p_j\right).
\end{equation}
It summarizes how evenly spectral energy is distributed. Cumulative spectral energy
at $k$ is $\sum_{j=1}^k p_j$ after sorting eigenvalues in descending order.
The calibration figure reports spectra of the raw second moment and the
mean-centered covariance, both computed before rowwise $\ell_2$ normalization.

\subsection{Error bands}
\label{app:error-bands}

Cosine-probe bands show the sample mean plus or minus $1.96$ standard errors.
Hit@5 bands use 95\% task-stratified example-bootstrap intervals from 5,000
resamples. These intervals quantify evaluation-example variation with the
models and probes held fixed. RSA curves show task-average point estimates.
For each task and depth in \cref{fig:task-depth-decodability}, we compute the
difference between the two models'
Hit@5 indicators on each shared example. We resample these example-wise
differences with replacement 5,000 times and take the 2.5th and 97.5th
percentiles of their mean. Dots mark intervals excluding zero. These are
pointwise intervals for individual task-depth comparisons.

\section{Datasets and Evaluation}
\label{app:integrity}

\subsection{Chemical tasks}

Our evaluation uses the following tasks from ChemCoTBench
\citep{hao2026beyond}.

\noindent\textbf{Molecular optimization.}
Given a source molecule and a requested property, the model generates a
modified molecule. Unlike instruction-specified editing, this task requires
choosing structural changes that improve the property. We consider LogP,
aqueous solubility, and quantitative estimation of drug-likeness (QED),
along with activity-related scores for DRD2, JNK3, and GSK3-$\beta$.
The evaluator uses TDC property oracles for LogP, QED, and the three activity
scores, and an ESOL estimator for solubility. These are computational scores
of the generated structures, rather than experimental measurements.

For property $p$, improvement is the difference
$\Delta_p=f_p(m_{\mathrm{generated}})-f_p(m_{\mathrm{source}})$.
Following the benchmark implementation, missing, invalid, or unscorable
outputs contribute zero improvement. Within each completion stream, the
evaluator clips improvements to their 5th and 95th percentiles before
computing the mean. Success rate is the fraction of outputs with
$\Delta_p>0$, computed before this clipping. We average the resulting metrics
over the three completion streams. Thus optimization measures improvement
relative to the source, not recovery of a unique reference molecule.

\noindent\textbf{Molecule editing.}
The input specifies a source molecule and an addition, deletion, or
substitution operation. The model must produce a molecule implementing that
instruction. We report operation correctness separately for the three
editing tasks.

\noindent\textbf{Molecular understanding and scaffold abstraction.}
Functional-group and ring-counting tasks ask the model to identify structural
features of an input molecule; we report mean absolute count error.
The Murcko task asks for the molecule's scaffold, the core formed by its
rings and connecting linkers. The evaluator extracts scaffolds from the
prediction and reference: identical scaffolds score one; otherwise it uses
Tanimoto similarity between radius-two, 1,024-bit Morgan fingerprints,
with zero assigned to invalid outputs or pairs without a common substructure.
The ring-system task asks whether a supplied ring structure occurs in the
input molecule and requires a yes/no answer; we report label accuracy.

\noindent\textbf{Forward-reaction prediction.}
Given the reaction inputs, the model predicts either the major product or
the byproduct. We report fingerprint Tanimoto similarity (FTS) to the
reference outcome and the fraction of chemically parseable outputs
(validity), separately for the two tasks.
FTS is the arithmetic mean of Tanimoto similarities computed using RDKit
topological, MACCS-key, and radius-two Morgan count fingerprints. These
encode paths, predefined structural features, and local atom neighborhoods,
respectively. Higher FTS indicates greater structural similarity on a
zero-to-one scale. This three-fingerprint score differs from the Morgan-only
comparison used for RSA in \cref{app:rsa-methods}. Outputs that RDKit cannot
parse as molecules receive zero similarity and remain in the denominator.
Validity measures parsability rather than correctness of the reaction outcome.

\subsection{Evaluation set and scoring}
\label{app:evaluation-scoring}

The evaluation contains 1,020 optimization, molecular understanding, scaffold,
and reaction examples, together with 100 editing examples. The tables report
six property-improvement scores, three editing accuracies, two count errors,
two scaffold scores, and four reaction metrics. Optimization success rates
appear in \cref{tab:optimization-success}.

We associate predictions with their reference examples using dataset
indices and example identifiers before scoring. We will release the
evaluation examples, canonicalization rules, and example-level outputs.
We use the standard ChemCoTBench test set and the decoding settings
specified in \cref{app:contract}.

\noindent\textbf{Sampling and aggregation.}
\label{app:complete-results}
For each checkpoint, we sample three responses per problem and score each
response separately. Let $S_j$ be the task score computed using the $j$th
response to every problem. We report $\frac{1}{3}\sum_{j=1}^{3}S_j$;
we do not select the best response. For editing accuracy, this is the
fraction of correct responses across all three samples, rather than the
fraction of problems solved by at least one sample.
Where reported, the sample standard deviation of $S_1,S_2,S_3$ describes
decoding variability for the same checkpoint. Intervals in the representation analyses summarize
variation across evaluation examples, as detailed in Appendix~\ref{app:analysis-methods}.

\section{Optimization Success Rates}
\label{app:optimization-success}

Success rate is the fraction of scored generations with positive property
improvement. We average over the three completion streams used in our
evaluations.

\begin{table}[t]
\centering
\small
\renewcommand{\arraystretch}{1.12}
\setlength{\tabcolsep}{4pt}
\caption{Molecular optimization success rates (\%). Dark-red bold values in
the \method{} row match or improve upon Coconut-Chem at the reported
precision; regressions remain black.}
\label{tab:optimization-success}
\begin{tabular}{lrrrrrr}
\toprule
Method & LogP & Solubility & QED & DRD2 & JNK3 & GSK3-$\beta$ \\
\midrule
LatentChem & 77.7 & 73.0 & 78.3 & 63.7 & 45.7 & 69.3 \\
\textbf{Coconut-Chem} & 82.3 & 85.0 & 89.0 & 77.3 & 58.3 & 81.3 \\
\rowcolor{tablemethodhighlight}
Coconut-Chem + \method{} & \localtie{82.3} & \localwin{86.0} & 86.3
& \localwin{79.3} & \localwin{64.0} & \localwin{84.0} \\
\bottomrule
\end{tabular}
\end{table}

\end{document}

%% file: math_commands.tex
\usepackage{amsmath,amsfonts,bm}

\def\eqref#1{equation~\ref{#1}}

\def\1{\bm{1}}

\DeclareMathAlphabet{\mathsfit}{\encodingdefault}{\sfdefault}{m}{sl}
\SetMathAlphabet{\mathsfit}{bold}{\encodingdefault}{\sfdefault}{bx}{n}



%% file: latentchem_transfer_appendix.tex
\section{Transfer to LatentChem}
\label{app:latentchem-transfer}

This experiment evaluates answer-representation prediction during supervised
adaptation, before reinforcement learning. Both arms share the same training
budget and trainable modules. \Cref{tab:latentchem-transfer} reports the complete
task results; the baseline is the same adapted LatentChem model reported in
\cref{tab:molopt-results,tab:structure-reaction-results}.

\begin{table*}[t]
\centering
\normalsize
\renewcommand{\arraystretch}{1.15}
\setlength{\tabcolsep}{10pt}
\caption{Answer-representation prediction with LatentChem. Both models receive
three epochs of supervised adaptation with the molecular updater frozen and
are evaluated at $T=0.7,p=0.9,n=3$. \method{} predicts an independently encoded
answer representation from the final latent thought. In the \method{} column,
dark-red bold indicates improvement over the adapted LatentChem baseline.
Editing accuracy, ring-system accuracy, and reaction validity are percentages;
optimization entries are mean property improvements, count entries are MAE,
and Murcko and FTS are similarities.}
\label{tab:latentchem-transfer}
\begin{tabular*}{\textwidth}{@{\extracolsep{\fill}}lrr@{}}
\toprule
Metric & LatentChem & LatentChem + \method{} \\
\midrule
\rowcolor{tablegrouplatent}
\multicolumn{3}{l}{\textit{Molecule editing (\%) $\uparrow$}} \\
Addition & 65.0 & 63.3 \\
Deletion & 78.3 & \localwin{85.0} \\
Substitution & 56.7 & \localwin{60.0} \\
\addlinespace
\rowcolor{tablegrouplatent}
\multicolumn{3}{l}{\textit{Molecular optimization: property improvement $\uparrow$}} \\
LogP & 0.7051 & \localwin{0.8040} \\
Solubility & 0.7829 & \localwin{0.8295} \\
QED & 0.1809 & \localwin{0.2053} \\
DRD2 & 0.1732 & \localwin{0.1907} \\
JNK3 & 0.0358 & \localwin{0.0500} \\
GSK3-$\beta$ & 0.1481 & 0.1320 \\
\addlinespace
\rowcolor{tablegrouplatent}
\multicolumn{3}{l}{\textit{Molecular understanding}} \\
Functional-group count error $\downarrow$ & 0.0967 & \localwin{0.0733} \\
Ring count error $\downarrow$ & 0.1667 & \localwin{0.1000} \\
Murcko scaffold similarity $\uparrow$ & 0.7220 & 0.6912 \\
Ring-system accuracy (\%) $\uparrow$ & 73.89 & 72.78 \\
\addlinespace
\rowcolor{tablegrouplatent}
\multicolumn{3}{l}{\textit{Reaction prediction}} \\
Major-product FTS $\uparrow$ & 0.5099 & \localwin{0.5541} \\
Major-product validity (\%) $\uparrow$ & 74.00 & \localwin{78.33} \\
Byproduct FTS $\uparrow$ & 0.1728 & \localwin{0.1735} \\
Byproduct validity (\%) $\uparrow$ & 42.67 & \localwin{45.33} \\
\bottomrule
\end{tabular*}
\end{table*}